\documentclass[twoside,11pt]{article}

\usepackage{blindtext}

\usepackage{jmlr2e}
\usepackage{algorithm}
\usepackage{algpseudocode}
\usepackage{multirow}
\RequirePackage{amsmath,amsfonts}

\newcommand{\deltain}{\delta_\text{in}}
\newcommand{\deltaout}{\delta_\text{out}}
\newcommand{\Din}{D^\text{in}}
\newcommand{\Dout}{D^\text{out}}
\newcommand{\indic}[1]{1_{\{ #1 \}}}
\DeclareMathOperator*{\argmin}{arg\,min}
\DeclareMathOperator*{\argmax}{arg\,max}
\newcommand{\PP}{\mathbb{P}}

\usepackage{lastpage}
\jmlrheading{23}{2022}{1-\pageref{LastPage}}{1/21; Revised 5/22}{9/22}{21-0000}{Daniel Cirkovic and Tiandong Wang}

\ShortHeadings{Networks with Heterogeneous Reciprocity}{Cirkovic and Wang}
\firstpageno{1}

\begin{document}

\title{Modeling Random Networks with Heterogeneous Reciprocity}

\author{\name Daniel Cirkovic \email cirkovd@stat.tamu.edu \\
       \addr Department of Statistics\\
       Texas A\&M University \\
       College Station, TX 77843, USA
       \AND
       \name Tiandong Wang \email td\_wang@fudan.edu.cn \\
       \addr Shanghai Center for Mathematical Sciences \\
       Fudan University \\
       Shanghai 200438, China}

\editor{My editor}

\maketitle

\begin{abstract}
Reciprocity, or the tendency of individuals to mirror behavior, is a key measure that describes information exchange in a social network. Users in social networks tend to engage in different levels of reciprocal behavior. Differences in such behavior may indicate the existence of communities that reciprocate links at varying rates. In this paper, we develop methodology to model the diverse reciprocal behavior in growing social networks. In particular, we present a preferential attachment model with heterogeneous reciprocity that imitates the attraction users have for popular users, plus the heterogeneous nature by which they reciprocate links. We compare Bayesian and frequentist model fitting techniques for large networks, as well as computationally efficient variational alternatives. Cases where the number of communities are known and unknown are both considered. We apply the presented methods to the analysis of a Facebook wallpost network where users have non-uniform reciprocal behavior patterns. The fitted model captures the heavy-tailed nature of the empirical degree distributions in the Facebook data and identifies multiple groups of users that differ in their tendency to reply to and receive responses to wallposts.
\end{abstract}


\section{Introduction}

A frequent goal in the statistical inference of social networks is to develop models that adequately capture and quantify common types of user interaction. One such feature is the propensity of users to generate links with other users that already have attracted a large number of links \citep{newman2001clustering, jeong2003measuring}. In order to model this ``rich get richer" self-organizing feature of nodes in a growing network, \cite{barabasi1999emergence} developed the preferential attachment (PA) model. The classical the preferential attachment model posits that as users enter a growing network, they connect with other users with probability proportional to their degree. This simple mechanism produces power-law degree distributions, yet another feature of many real-world networks \citep{mislove2007measurement}. Since it's inception, many generalizations of the preferential attachment model have been developed to capture more features of growing networks \citep{bhamidi2015twitter, hajek2019community, wang2022directed, wang2020directed}.

Another common feature of online social networks is a significant degree of reciprocity \citep[see][for example]{newman2002email, zlatic2011model}. Reciprocity describes the tendency of users to reply to links and is typically measured by the proportion of reciprocal links in a network \citep{jiang2015reciprocity}. A recent study by \cite{wang2021measuring} found that the traditional directed preferential attachment model often produces a negligible proportion of reciprocal links. Motivated by this finding, \cite{wang2022asymptotic} and \cite{cirkovic2022preferential} developed a preferential attachment model with reciprocity that is a more realistic choice for fitting to social networks. The model assumes that upon the generation of a link between nodes through the typical preferential attachment scheme, the users reciprocate the link with a probability $\rho \in (0, 1)$ that is common to all users in the network. The model was used to analyze a Facebook wallpost network.

Although an improvement, the model of \cite{cirkovic2022preferential} fails to account for the heterogeneity of reciprocal behavior in a social network. In reality, it is na\"ive to assume all users in a large network engage in similar levels of reciprocity. Such an assumption has caused \cite{cirkovic2022preferential} to remove a subset of nodes which apparently engaged in dissimilar reciprocal behavior from their analysis of the Facebook wallpost network. Further, when a link is made between two nodes $u$ and $v$, it is likely that the decision of whether or not to reciprocate the link depends on the direction of the original link, $(u,v)$ or $(v, u)$. For example, a celebrity in a social network may be less likely to reply to a message sent by a fan, whereas a fan is very likely to respond to message sent by the celebrity. Recently, \cite{wang2022random} relax the assumption of having only one reciprocity parameter $\rho$ to the case where reciprocity probabilities are different for users belonging to different communication classes. Theoretical results in \cite{wang2022asymptotic} are obtained by assuming no new edge is added between existing nodes.

In this paper, we consider a further generalization of the model presented in \cite{wang2022random} to allow for more realistic assumptions, i.e. heterogeneous, asymmetric reciprocity as well as edges between existing nodes. We assume that each user in the network is equipped with a communication class that governs its tendency to reciprocate edges. In the network generation process, initial edges between nodes are generated via preferential attachment, while the decision to reciprocate the edge is decided by a stochastic blockmodel-like scheme. We describe three methods to fit such a model to observed networks, both when the number of communication classes is known and unknown. Specifically, we propose a fully Bayesian approach, along with variationally Bayesian and frequentist approaches. The approaches and their peformance on synthetic networks are then compared through simulation studies. Finally, we reconsider the Facebook wallpost network as in \cite{cirkovic2022preferential}, and use the heterogeneous reciprocal preferential attachment model to glean new insights into communication patterns on Facebook.

\section{The PA Model with Heterogeneous Reciprocity}

\subsection{The model}\label{sec:PAmodel}

In this section, we present the preferential attachment model with heterogeneous reciprocity. Let $G(n)$ be the graph after $n$ steps and $V(n)$ be the
set of nodes in $G(n)$.  
Attach to each node $v$ a communication type $W_v$, where $\{W_v, v\geq 1\}$ are iid random variables with
\begin{equation}
\PP(W_v = r) =\pi_r,\qquad\text{for}\quad \sum_{r=1}^K \pi_r = 1.
\label{eq:pi}
\end{equation}
Define the vector $\boldsymbol{\pi} \equiv (\pi_r)_r$. Let $W(n) := \{W_v: v\in V(n)\}$ denote the set of group types for all nodes in $G(n)$.
Throughout we assume that the communication group of node $v$ is generated upon creation and remains unchanged throughout the graph evolution.
Also, denote the set of directed edges in $G(n)$ by
\[
E(n) := \{(u,v): u,v\in V(n)\}.
\]
Throughout this paper, we always assume $G(n) = (V(n), E(n), W(n))$, for $n\ge 0$.

We initialize the model with seed graph $G(0)$. $G(0)$ consists of $|V(0)|$ nodes, each of which is also endowed with its own communication class randomly according to \eqref{eq:pi}.
The edges $E(0)$ will have no impact on inference other than setting the initial degree distribution.  
For each new edge $(u, v)$ with $W_u = r, W_v=m$, the reciprocity mechanism adds its reciprocal
counterpart $(v, u)$ instantaneously with probability $\rho_{m, r}\in [0,1]$, for $m,r\in \{1,2,\ldots,K\}$. 
Here $\rho_{m, r}$ measures the probability of adding a reciprocal edge from a node in group $m$ to a node in group $r$.
Note that the matrix $\boldsymbol{\rho} := (\rho_{m,r})_{m,r}$ is not necessarily a stochastic matrix, but can be an arbitrary matrix in $M_{K\times K}([0,1])$, the set of all $K\times K$ matrices with entries belonging to $[0,1]$.

We now describe the evolution of the network $G(n+1)$ from $G(n)$.
Let $\bigl(\Din_v(n), \Dout_v(n)\bigr)$
be the in- and out-degrees of node $v\in V(n)$, and 
we use the convention that $\Din_v(n) = \Dout_v(n) = 0$ if
$v\notin V(n)$. 
\begin{enumerate}
\item With probability $\alpha \in [0, 1]$, 
add a new node $|V(n)|+1$ with a directed edge $(|V(n)|+1,v)$,
where $v\in V(n)$ is chosen with probability
\begin{equation}\label{eq:alpha_prob}
\frac{\Din_v(n)+\deltain}{\sum_{v\in V(n)} (\Din_w(n)+\deltain)}
= \frac{\Din_v(n)+\deltain}{|E(n)|+\deltain |V(n)|},
\end{equation}
where $\deltain > 0$ is an offset parameter,
and update the node set $V(n+1)=V(n)\cup\{|V(n)|+1\}$ and 
$W(n+1)=W(n)\cup\{W_{|V(n)|+1}\}$.
The new node $|V(n)|+1$ belongs to group $r$ with probability $\pi_r$.
If node $v$ belongs to group $m$, then 
a reciprocal edge $(v, |V(n)|+1)$ is added
with probability $\rho_{m, r}$. Update the edge set as
$E(n+1) = E(n)\cup \{(|V(n)|+1,v), (v,|V(n)|+1)\}$.
If the reciprocal edge is not created, set $E(n+1) = E(n)\cup \{(|V(n)|+1,v)\}$.

\item With probability $\beta \in [0, 1 - \alpha]$, generate a directed edge $(u, v)$ between two existing nodes $u, v \in V(n)$ with probability
\begin{equation}\label{eq:beta_prob}
\begin{split}
\frac{\Din_v(n)+\deltain}{\sum_{v\in V(n)} (\Din_w(n)+\deltain)}&\frac{\Dout_v(n)+\deltaout}{\sum_{v\in V(n)} (\Dout_v(n)+\deltaout)} \\
&= \frac{\Din_v(n)+\deltain}{|E(n)|+\deltain |V(n)|} \frac{\Dout_v(n)+\deltaout}{|E(n)|+\deltaout |V(n)|},
\end{split}
\end{equation}
where $\deltaout > 0$ is also an offset parameter. 
If node $u$ belongs to group $r$ and node $v$ belongs to group $m$, then a reciprocal edge $(v, u)$ is added with probability $\rho_{m ,r}$. 
Update the edge set as $E(n+1) = E(n)\cup \{(u, v), (v, u)\}$.
If the reciprocal edge is not created, set $E(n+1) = E(n)\cup \{(u,v)\}$.
Finally, update $V(n+1)=V(n)$ and $W(n+1)=W(n)$. 

\item With probability $\gamma \equiv 1 - \alpha - \beta$, 
add a new node $|V(n)|+1$ with a directed edge $(v,|V(n)|+1)$,
where $v\in V(n)$ is chosen with probability
\begin{equation}\label{eq:gamma_prob}
\frac{\Dout_v(n)+\deltaout}{\sum_{v\in V(n)} (\Dout_v(n)+\deltaout)}
= \frac{\Dout_v(n)+\deltaout}{|E(n)|+\deltaout |V(n)|},
\end{equation}
 and update the node set $V(n+1)=V(n)\cup\{|V(n)|+1\}$, 
$W(n+1)=W(n)\cup\{W_{|V(n)|+1}\}$.
The new node $|V(n)|+1$ belongs to group $r$ with probability $\pi_r$.
If node $v$ belongs to group $m$, then 
a reciprocal edge $(|V(n)|+1,v)$ is added
with probability $\rho_{r, m}$.
Update the edge set as 
$E(n+1) = E(n)\cup \{(v, |V(n)|+1,v), (|V(n)|+1,v)\}$.
If the reciprocal edge is not created, set $E(n+1) = E(n)\cup
\{(v,|V(n)|+1)\}$.
\end{enumerate}

Let $\{J_k\}$ be iid Categorical random variables that indicate under which scenario the transition from $G(k)$ to $G(k + 1)$ has occurred. That is, $\mathbb{P}(J_k = 1) = \alpha$, $\mathbb{P}(J_k = 2) = \beta$ and $\mathbb{P}(J_k = 3) = 1 - \alpha - \beta$. 
At each step $k$, we denote the outcome of the reciprocal event via $R_k$ where $R_k =1$ if a reciprocal edge is added and $R_k = 0$ otherwise. 

\subsection{Likelihood inference}\label{sec:likelihood}

Suppose we observe the evolution of the graph sequence $\{ G(k) \}_{k = 0}^n$ so that we have the edges $e_k = E(k)\setminus E(k - 1)$ added at each step according to the description in Section \ref{sec:PAmodel}. Here,
\begin{equation}
e_k = 
\begin{cases}
{\{(s_k, t_k), (t_k, s_k)} \} &\text{if } R_k = 1 \\
\{(s_k, t_k)\} &\text{if } R_k = 0.
\end{cases}
\end{equation}
Let $\boldsymbol{\theta} = (\alpha, \beta, \deltain, \deltaout)$. With these ingredients, the likelihood associated with the graph sequence $\{ G(k) \}_{k = 0}^n$ is given by
\begin{align*}
& p \left((e_k)_{k = 1}^{n}, W(n) \mid  \boldsymbol{\theta}, \boldsymbol{\pi}, \boldsymbol{\rho}  \right) \\
& = \alpha^{\sum_{k = 1}^n\indic{J_k = 1}} \beta^{\sum_{k = 1}^n\indic{J_k = 2}} (1 - \alpha - \beta)^{\sum_{k = 1}^n\indic{J_k = 3}} \\
& \hspace{0.5cm} \times \prod_{k = 1}^n  \left( \frac{\Din_{t_k}(k - 1) + \deltain}{|E(k-1)| + \deltain |V(k - 1)|} \right)^{\indic{J_k \in \{1, 2 \}}} \left(\frac{\Dout_{s_k}(k - 1) + \deltaout}{|E(k-1)| + \deltaout |V(k - 1)|} \right)^{\indic{J_k \in \{2, 3\}}} \\ 
& \hspace{1cm} \times \prod_{r = 1}^K \pi^{\sum_{k = 1}^n \indic{J_k = 1} \indic{W_{s_k = r}} + \sum_{k = 1}^n \indic{J_k = 3} \indic{W_{t_k = r}}}_r \\
& \hspace{1.5cm} \times \prod_{r = 1}^K \prod_{m = 1}^K \rho^{\sum_{k = 1}^n \indic{W_{s_k} = r}\indic{W_{t_k} = m}\indic{R_k = 1}}_{m, r} (1 - \rho_{m, r})^{\sum_{k = 1}^n \indic{W_{s_k} = r}\indic{W_{t_k} = m}\indic{R_k = 0}} \\
&\equiv p((e_k)_{k = 1}^{n} \mid \boldsymbol{\theta}) \times p((e_k)_{k = 1}^{n}, W(n) \mid \boldsymbol{\pi}, \boldsymbol{\rho}).
\end{align*}

The function $p(\cdot \mid \boldsymbol{\theta})$ collects the likelihood terms dependent on $\boldsymbol{\theta}$ and likewise $p( \cdot \mid \boldsymbol{\pi}, \boldsymbol{\rho})$ collects the terms dependent on $\boldsymbol{\pi}$ and $\boldsymbol{\rho}$. 
Such factorization implies that the estimation of the parameters $\boldsymbol{\theta}$ and $\boldsymbol{\pi}, \boldsymbol{\rho}$ can be conducted independently. 
The frequentist estimation of $\boldsymbol{\theta}$ in homogeneous reciprocal PA models has already been considered in \cite{cirkovic2022preferential}. 
These estimators are unchanged in the heterogeneous case. Naturally, the maximum likelihood estimators (MLE) for $\alpha$ and $\beta$ are given by $\hat{\alpha} = n^{-1}\sum_{k = 0}^n \indic{J_k = 1}$ and $\hat{\beta} = n^{-1}\sum_{k = 0}^n \indic{J_k = 2}$. The MLE for $\deltain$ satisfies
\begin{equation}
\label{eq:deltaMLE}
\sum_{k = 1}^n \indic{J_k \in \{1, 2 \}} \frac{1}{\Din_{t_k}(k - 1) + \hat{\delta}_\text{in}} - \sum_{k = 1}^n \indic{J_k \in \{1, 2 \}} \frac{|V(k - 1)|}{|E(k - 1)| + \hat{\delta}_\text{in} N(k - 1)} = 0,
\end{equation}
where \eqref{eq:deltaMLE} is obtained by setting $\frac{\partial}{\partial \deltain} \log p((e_k)_{k = 1}^{n} \mid \boldsymbol{\theta}) = 0$. 
The MLE for $\deltaout$ is obtained similarly. 
The estimators $\hat{\alpha}$ and $\hat{\beta}$ are strongly consistent for $\alpha$ and $\beta$, while consistency for $\hat{\delta}_\text{in}$ and $\hat{\delta}_\text{out}$ has not yet been verified since the reciprocal component of the model interferes with traditional techniques to analyze consistency in non-reciprocal preferential attachment models as in \cite{wan2017fitting}. Estimation of $\boldsymbol{\rho}$ and $\boldsymbol{\pi}$ is considerably more involved, and will be the main focus of this paper. 

The reciprocal component of the preferential attachment model with heterogeneous reciprocity is reminiscent of a stochastic block model. Nodes first attach via the preferential attachment rules in \eqref{eq:alpha_prob}, \eqref{eq:beta_prob} and \eqref{eq:gamma_prob}, then a stochastic-block-model type mechanism dictates the reciprocal behavior. A large portion of the literature on stochastic block modeling is concerned with community detection \cite{bickel2009nonparametric, holland1983stochastic, karrer2011stochastic, zhao2011community}. Here we are primarily concerned with the estimation of $\boldsymbol{\rho}$ and $\boldsymbol{\pi}$, and consider the recovery of $W(n)$ as a secondary goal. The optimal recovery of $\boldsymbol{\rho}$ and $\boldsymbol{\pi}$ hinges on the correct specification of $K$, the number of reciprocal clusters. We will thus examine cases when $K$ is known a priori, as well as cases where it must be inferred from the data. 

We also note that a minor nuisance of modeling reciprocal PA models is the observation of the random variable $R_k$. Upon observations the edges $\{(s_k, t_k), (t_k, s_k)\}$, it is not possible to identify whether the second edge was generated under $R_k = 1$ or $J_k = 2$. Since, upon observation, the probability that the edge was generated under $J_k = 2$ is extremely small for large networks, we assume all such reciprocated edges are generated under $R_k = 1$. In real-world networks, however, time will often pass between message replies. For such networks, we will thus employ window estimators from \cite{cirkovic2022preferential}. We defer further discussion of window estimators to Section \ref{sec:Facebook}.

We will continue to consider the estimation of $\boldsymbol{\rho}$ and $\boldsymbol{\pi}$ based on $p((e_k)_{k = 1}^{n}, W(n) \mid \boldsymbol{\pi}, \boldsymbol{\rho})$. Since $W(n)$ is unobservable, a natural probabilistic approach would marginalize over the unobservable communication types, and form a complete-data likelihood $p((e_k)_{k = 1}^{n} \mid \boldsymbol{\pi}, \boldsymbol{\rho})$. This, however, involves a sum over all latent configurations of $W(n)$ which is analytically intractable, as well as computationally infeasible for large networks. Such difficulties encourage attempts to learn $W(n)$ from the conditional distribution of $W(n)$ given $(e_k)_{k = 1}^{n}$ (\'{a} la an EM Algorithm \cite{dempster1977maximum}) and jointly estimate $W(n), \boldsymbol{\pi}$ and $\boldsymbol{\rho}$. Often, these attempts are computationally infeasible due to the lack of factorization in the conditional distribution. In the following section, we will consider both Bayesian and frequentist estimation methods for $\boldsymbol{\pi}$ and $\boldsymbol{\rho}$ where $K$ is known. We will first present an ``ideal" fully Bayesian approach, and then move on to variationally Bayesian and frequentist approximations to that ideal. Afterwards, we will discuss how to perform model selection when $K$ is unknown for each of these methods.

\section{Inference for a known number of communication types}\label{sec:knownK}

\subsection{Bayesian inference}\label{sec:BayesKknown}

For Bayesian inference of the heterogeneous reciprocal PA model we follow \cite{nowicki2001estimation} and employ independent and conditionally conjugate priors
\begin{align}
\label{eq:priors}
\begin{split}
&\rho_{m, r} \overset{\text{i.i.d.}}{\sim} \text{Beta}(a, b), \ m, r = 1, \dots, K, \\
& \boldsymbol{\pi} \sim \text{Dirichlet}(\eta, \dots, \eta).
\end{split}
\end{align}
The prior specification \eqref{eq:priors} leads to a simple Gibbs sampler that draws approximate samples from the posterior $p \left( \boldsymbol{\rho}, \boldsymbol{\pi}, W(n) \mid (e_k)_{k = 1}^n \right)$. We present the Gibbs sampler as Algorithm \ref{alg:Gibbs}. Here, $\boldsymbol{\rho}$ and $\boldsymbol{\pi}$ are initialized from prior draws and $W(n)$ is initialized by drawing from $p\left(W_v \mid \boldsymbol{\pi} \right)$ for $v = 1, \dots |V(n)|$. Although the sampler is standard, many samples are required to sufficiently explore the posterior distribution. For large networks, this can be computationally onerous, and hence we appeal to variational alternatives.

\begin{algorithm}
\caption{Gibbs sampling for heterogeneous reciprocal PA with known $K$}
\label{alg:Gibbs}
\begin{algorithmic}
\Require Graph $G(n)$, \# communication types $K$, prior parameters $a$, $b$, $\eta$, \# MCMC iterations $M$
\Ensure Approximate samples from the posterior $p \left( \boldsymbol{\rho}, \boldsymbol{\pi}, W(n) \mid (e_k)_{k = 1}^n \right)$ \\
\textbf{Initialize:} Draw $\boldsymbol{\pi}$ and $\boldsymbol{\rho}$ from \eqref{eq:priors}, draw $W_v \sim \text{Multinomial}(\boldsymbol{\pi})$ for $v \in V(n)$
\For{$i = 1$ to $M$}
\Statex 1. Sample $W(n)$ from its conditional posterior
\ForAll {$v \in V(n)$}
\State Sample $W_v$ according to 
\begin{align*}
P&\left( W_v = \ell \mid  \boldsymbol{\pi}, \boldsymbol{\rho}, (W_u)_{u \neq v}, (e_k)_{k = 1}^n  \right) \\
&\propto \pi_r\prod_{m = 1}^K  \rho_{m, \ell}^{\sum_{k:s_k = v} 1_{\{W_{t_k} = m \}}1_{\{ R_k = 1 \}}} 
  (1 - \rho_{m, \ell})^{\sum_{k:s_k = v} 1_{\{W_{t_k} = m \}} 1_{\{ R_k = 0 \}}} \\
&\qquad\times\prod_{r = 1}^K  \rho_{\ell, r}^{\sum_{k:t_k = v} 1_{\{W_{s_k} = r \}}1_{\{ R_k = 1 \}}} 
(1 - \rho_{\ell, r})^{\sum_{k:t_k = v} 1_{\{W_{s_k} = r \}} 1_{\{ R_k = 0 \}}}
\end{align*}
\State for $\ell = 1,\dots, K$
\EndFor 
\Statex 2. Sample $\boldsymbol{\rho}$ from its conditional posterior 
\For{$m = 1$ to $K$}
\For{$r = 1$ to $K$}
\State Sample $\rho_{m ,r}$ from
\begin{align*}
\rho_{m ,r} \mid \boldsymbol{\pi}, W(n), (e_k)_{k = 1}^n  \sim \text{Beta}\Bigg(a &+ \sum_{k = 1}^n 1_{\{W_{s_k = r}\}}1_{\{W_{t_k = m} \}}1_{\{R_k = 1\}}, \\
&b + \sum_{k = 1}^n 1_{\{W_{s_k = r}\}}1_{\{W_{t_k = m} \}}1_{\{R_k = 0\}} \Bigg)
\end{align*}
\EndFor
\EndFor
\Statex 3. Sample $\boldsymbol{\pi}$ from its conditional posterior 
\begin{align*}
\boldsymbol{\pi} \mid \boldsymbol{\rho},  W(n), (e_k)_{k = 1}^n \sim \text{Dirichlet}\left( \eta + \sum_{v \in V(n)} 1_{\{W_v = 1 \}}, \dots, \eta + \sum_{v \in V(n)} 1_{\{W_v = K \}} \right)
\end{align*}
\EndFor
\end{algorithmic}
\end{algorithm}

\subsection{Variational inference}\label{sec:Variational}

In this section, we present variational alternatives for approximating posteriors associated with the heterogeneous reciprocal PA model. The aim of variational inference is to approximate the conditional distribution of latent variables $\mathbf{z}$ given data $\mathbf{x}$ via a class of densities $\mathcal{Q}$ typically chosen to circumvent computational inconveniences. If Bayesian inference is being performed, the latent variables $\mathbf{z}$ can also encompass the model parameters ($\boldsymbol{\pi}$ and $\boldsymbol{\rho}$ in our setting). The variational inference procedure aims to find the density $q^\star \in \mathcal{Q}$ that minimizes the Kullback-Leibler (KL) divergence from $p(\cdot \mid \mathbf{x})$, i.e.
\begin{align}
\label{eq:obj}
q^\star = \argmin_{q \in \mathcal{Q}} \text{KL}\left(q(\cdot) \mid\mid p(\cdot \mid \mathbf{x}) \right).
\end{align}
We will restrict $\mathcal{Q}$ to the mean-field family, that is, the family of densities where components $\mathbf{z}$ are mutually independent. Naturally, such restriction will prevent $q^\star$ from capturing the dependence structure between the latent variables. Recently, however, some more structured, expressive families have been proposed that may improve the approximation; see for instance \cite{yin2020theoretical}. Conveniently, using the definition of the conditional density, the objective \eqref{eq:obj} can be expressed as
\begin{align}
\label{eq:KL}
\text{KL}\left(q(\cdot) \mid\mid p(\cdot \mid \mathbf{x}) \right) = E_q[\log q(\mathbf{z})] - E_q[\log p(\mathbf{z}, \mathbf{x})] + \log p(\mathbf{x}) \equiv -\text{ELBO}(q) + \log p(\mathbf{x}),
\end{align}
so that minimizing the KL divergence from $p(\cdot \mid \mathbf{x})$ to $q(\cdot)$ is equivalent to maximizing the evidence lower bound ($\text{ELBO}(q)$) since $\log p(\mathbf{x})$ does not depend on $q$. For more on variational inference, see \cite{blei2017variational}.

\subsubsection{Bayesian Variational Inference}\label{sec:BVI}

Now we consider solving the variational problem \eqref{eq:obj} for the probabilistic model presented in Section \ref{sec:BayesKknown}. Although we have presented a sampler in Algorithm \ref{alg:Gibbs} that draws approximate samples from the posterior, we aim for an estimate that sacrifices modeling the dependence in the posterior distribution in favor of computation time. Variational inference for stochastic blockmodels in the Bayesian setting was studied in \cite{latouche2012variational}. Following their strategy, we posit a mean-field variational family:
\begin{align}
\label{eq:MFBayes}
q(\boldsymbol{\pi}, \boldsymbol{\rho}, W(n)) = q(\boldsymbol{\pi})q(\boldsymbol{\rho})q(W(n)) = q(\boldsymbol{\pi})\prod_{m = 1}^K\prod_{r = 1}^Kq(\rho_{m , r})\prod_{v \in V(n)}q_v(W_v).
\end{align}
We further assume that the variational densities have the following forms:
\begin{align*}
q(\boldsymbol{\pi}) &\propto \prod_{r = 1}^K \pi^{d_1}_1 \cdots \pi^{d_K}_K, \ d_1, \dots, d_K \geq 0, \\
q(\rho_{m, r}) &\propto \rho_{m , r}^{\omega_{m, r}}(1 - \rho_{m, r})^{\xi_{m, r}}, \ \omega_{m, r}, \xi_{m, r} \geq 0, \ m,r = 1,\dots, K, \\
q_v(W_v) &= \prod_{r = 1}^K \tau_{v, r}^{1_{\{W_v = r \}}}, \tau_{v, r} \geq 0, \ r = 1, \dots, K, \ v = 1, \dots |V(n)|,
\end{align*}
and additionally $\sum_{r = 1}^K \tau_{v, r} = 1$ for all $v \in V(n)$. In other words, the posterior of $\boldsymbol{\pi}$ is approximated by a $\text{Dirichlet}(d_1, \dots, d_K)$ distribution, and the component-wise posteriors of $\boldsymbol{\rho}$ and $W(n)$ are approximated by $\text{Beta}(\omega_{m, r}, \xi_{m, r})$ and $\text{Multinomial}(1, (\tau_{v, r})_{r = 1}^K)$ distributions, respectively. In Algorithm \ref{alg:CAVI} we present a coordinate ascent variational inference (CAVI) algorithm for optimizing the ELBO. Here, $\psi(\cdot)$ is the digamma function. Note that in step 3 of algorithm, we write $\sum_{k:s_k = v} \equiv \sum_{k:s_k = v, s_k \neq t_k}$ for brevity of notation. The inclusion of self-loops makes the optimization of the ELBO much more difficult, hence their exclusion. Here, the class probabilities, $\tau_{u, r}$, are initialized uniformly at random. We omit the calculations for the derivation of this algorithm, as they are very similar to \cite{latouche2012variational}.

\begin{algorithm}
\caption{CAVI for heterogeneous reciprocal PA with known $K$}
\label{alg:CAVI}
\begin{algorithmic}
\Require Graph $G(n)$, \# communication types $K$, prior parameters $a$, $b$, $\eta$, tolerance $\epsilon > 0$
\Ensure Variational approximation to the posterior $q^\star$ \\
\textbf{Initialize:} Draw $\tau_{v, r}$, $r = 1, \dots, K$ uniformly at random from the $K$-simplex for every $v \in V(n)$
\While{the increase in $\text{ELBO}(q)$ is greater than $\epsilon$}
\Statex 1. Update $q(\boldsymbol{\pi})$
\For{$r = 1$ to $K$}
\begin{align*}
d_r = \eta + \sum_{v \in V(n)} \tau_{u, r}
\end{align*}
\EndFor
\Statex 2. Update $q(\boldsymbol{\rho})$
\For{$m = 1$ to $K$}
\For{$r = 1$ to $K$}
\begin{align*}
\omega_{m, r} &= a + \sum_{k = 1}^n \tau_{s_k, r}\tau_{t_k, m} 1_{\{ R_k = 1 \}} \\ 
\xi_{m, r} &= b + \sum_{k = 1}^n \tau_{s_k, r}\tau_{t_k, m} 1_{\{ R_k = 0 \}}
\end{align*}
\EndFor
\EndFor
\Statex 3. Update $\text{ELBO}(q)$ according to \eqref{eq:ELBOVB}
\Statex 4. Update $q(W(n))$
\ForAll{$v \in V(n)$}
\For{$\ell = 1$ to $K$}
\begin{alignat*}{2}
\tau_{v, \ell} \propto & \exp \Bigg\lbrace \psi \left(d_\ell \right) && - \psi\Bigg(\sum_{r = 1}^K d_r\Bigg) \Bigg\rbrace \\
& \times \prod_{m = 1}^K \exp \Bigg\lbrace && \psi \left( \omega_{m, \ell} \right) \sum_{k: s_k = v }\tau_{t_k, m}1_{\{R_k = 1\}} + \psi \left( \xi_{m, \ell} \right) \sum_{k: s_k = v }\tau_{t_k, m}1_{\{R_k = 0\}} \\
& && - \psi(\omega_{m, \ell} + \xi_{m, \ell}) \sum_{k : s_k = v} \tau_{t_k, m}  \Bigg\rbrace \\
& \times \prod_{r = 1}^K \exp \Bigg\lbrace && \psi \left( \omega_{\ell, r} \right) \sum_{k: t_k = v }\tau_{s_k, r}1_{\{R_k = 1\}} + \psi \left( \xi_{\ell, r} \right) \sum_{k: t_k = v }\tau_{s_k, r}1_{\{R_k = 0\}} \\
& && - \psi(\omega_{\ell, r} + \xi_{\ell, r}) \sum_{k : t_k = v} \tau_{s_k, r}  \Bigg\rbrace
\end{alignat*}
\EndFor
\EndFor
\EndWhile
\end{algorithmic}
\end{algorithm}

To monitor the convergence of Algorithm \ref{alg:CAVI}, we recommend computing the ELBO after each iteration of the CAVI algorithm and terminating the algorithm once the increase in the ELBO is less than some predetermined threshold $\epsilon$. Specifically, if the ELBO is computed after step 2, it has the simplified form:
\begin{equation}
\label{eq:ELBOVB}
\begin{split}
\text{ELBO}(q) =& \log \left( \frac{\Gamma(K \eta) \prod_{r = 1}^K \Gamma(d_r)}{\Gamma(\sum_{r = 1}^K d_r) \Gamma(\eta)^K} \right) + \sum_{r = 1}^K \sum_{m = 1}^K \log \left( \frac{\Gamma(a + b) \Gamma(\omega_{m, r}) \Gamma(\xi_{m, r})}{\Gamma(\omega_{m, r} + \xi_{m, r})\Gamma(a)\Gamma(b)} \right) \\
& - \sum_{v \in V(n)}\sum_{r = 1}^K \tau_{v, r} \log \tau_{v, r}.
\end{split}
\end{equation}

\subsubsection{Variational Expectation Maximization}\label{sec:VEM}

In this section we consider frequentist estimation of the PA model with heterogeneous reciprocity through a variational expectation maximization algorithm (VEM). VEM for stochastic blockmodel data was first considered in \cite{daudin2008mixture} which further inspired many interesting generalizations that could enhance the reciprocal PA model \citep[see][for example]{matias2017statistical}. The VEM algorithm augments the traditional EM algorithm by approximating the E-step for models in which the conditional distribution of the latent variables given the observed data is computationally intractable. The VEM estimates thus serve as a computationally efficient approximation to the maximum likelihood estimates of $\boldsymbol{\pi}$ and $\boldsymbol{\rho}$. Although a frequentist procedure, the VEM algorithm may enhance Bayesian inference of stochastic blockstructure data. For example, since the dimension of the posterior $p\left(\boldsymbol{\pi}, \boldsymbol{\rho} \mid (e_k)_{k = 1}^n \right)$ does not grow with the size of the data, one might expect a Bernstein-von-Mises phenomena to occur. The VEM estimates may thus approximate the posterior mean or even be leveraged to enhance posterior sampling as in \cite{donnet2021accelerating}. 

As in Section \ref{sec:BVI}, we approximate the distribution of the communication types given the observed network,  $p\left( W(n) \mid \boldsymbol{\pi}, \boldsymbol{\rho}, (e_k)_{k = 1}^n \right)$, via the mean-field approximation
\begin{align*}
q(W(n)) = \prod_{v \in V(n)} q_v(W_v).
\end{align*} 
Via the mean-field family assumption, the ELBO is given by
\begin{align}
\label{eq:ELBOVEM}
\begin{split}
\text{ELBO}&(q, \boldsymbol{\pi}, \boldsymbol{\rho}) \\
=& E_q \left[ \log p\left( W(n), (e_k)_{k = 1}^n \mid \boldsymbol{\pi}, \boldsymbol{\rho} \right) \right] - E_q\left[\log q(W(n))\right] \\
=& \sum_{k = 1}^n \sum_{r = 1}^K \left( \indic{J_k = 1} \tau_{s_k, r} + \indic{J_k = 3} \tau_{t_k, r} \right) \log \pi_{r} - \sum_{v \in V(n)} \sum_{r = 1}^K \tau_{v, r} \log \tau_{v, r}   \\
& + \sum_{k = 1}^n \sum_{r = 1}^K \sum_{m = 1}^K \tau_{s_k, r} \tau_{t_k, m} \left( \indic{R_k = 1}  \log \rho_{m, r} + \indic{R_k = 0} \log( 1-  \rho_{m, r}) \right).
\end{split}
\end{align}
Note that from \eqref{eq:KL}, maximizing \eqref{eq:ELBOVEM} with respect to $q$ (the E-step) is equivalent to minimizing the KL divergence from $p\left( \cdot \mid \boldsymbol{\pi}, \boldsymbol{\rho}, (e_k)_{k = 1}^n \right)$ to $q(\cdot)$ and maximizing \eqref{eq:ELBOVEM} with respect to $\boldsymbol{\pi}$ and $\boldsymbol{\rho}$ is equivalent to the M-step in the usual EM algorithm. Thus, the E-step is equivalent to performing variational inference for $p\left( \cdot \mid \boldsymbol{\pi}, \boldsymbol{\rho}, (e_k)_{k = 1}^n \right)$ where $\boldsymbol{\pi}$ and $\boldsymbol{\rho}$ are evaluated at their current estimates $\hat{\boldsymbol{\pi}}_\text{VEM}$ and $\hat{\boldsymbol{\rho}}_\text{VEM}$.

The VEM algorithm for the heterogeneous reciprocal PA model is given in Algorithm \ref{alg:VEM}. As in Algorithm \ref{alg:CAVI}, we write $\sum_{k:s_k = v} \equiv \sum_{k:s_k = v, s_k \neq t_k}$ for ease of notation. We describe the intialization of the algorithm at the end of Appendix A in Algorithm \ref{alg:VEMinit}. We further provide some derivations of the VEM algorithm in Appendix B. Similar types of computations can be employed to derive Algorithm \ref{alg:CAVI}. As in Algorithm \ref{alg:CAVI}, we recommend cycling through the updates of $\hat{\tau}_{v, \ell}$ in the E-step until the ELBO no longer increases beyond a prespecified threshold $\epsilon > 0$. 

\begin{algorithm}
\caption{VEM for heterogeneous reciprocal PA with known $K$}
\label{alg:VEM}
\begin{algorithmic}
\Require Graph $G(n)$, \# communication types $K$, tolerances $\epsilon, \kappa > 0$
\Ensure Variational EM estimates $\hat{\boldsymbol{\pi}}_\text{VEM}$ and $\hat{\boldsymbol{\rho}}_\text{VEM}$ \\
\textbf{Initialize:} Draw $\hat{\tau}_{v, r}$, $r = 1, \dots, K$ uniformly at random from the $K$-simplex for every $v \in V(n)$, run Algorithm \ref{alg:VEMinit} to initialize $\hat{\boldsymbol{\pi}}_\text{VEM}$ and $\hat{\boldsymbol{\rho}}_\text{VEM}$
\While{at least one of the elements of $\hat{\boldsymbol{\pi}}_\text{VEM}$ and $\hat{\boldsymbol{\rho}}_\text{VEM}$ change by more than $\kappa$ in absolute value}
\Statex 1. \textbf{E-step}: Update $\hat{q}$ via
\While{the increase in $\text{ELBO}(q)$ is greater than $\epsilon$}
\ForAll{$v \in V(n)$}
\For{$\ell = 1$ to $K$}
\begin{align*}
\hat{\tau}_{v, \ell} \propto \hat{\pi}_\ell \prod_{m = 1}^K & \hat{\rho}_{m , \ell}^{\sum_{k: s_k = v} \hat{\tau}_{t_k, m} \indic{R_k = 1}} (1 - \hat{\rho}_{m , \ell})^{\sum_{k: s_k = v} \hat{\tau}_{t_k, m} \indic{R_k = 0}} \\
&\times \prod_{r = 1}^K \hat{\rho}_{\ell, r}^{\sum_{k: t_k = v} \hat{\tau}_{s_k, r} \indic{R_k = 1}} (1 - \hat{\rho}_{\ell , r})^{\sum_{k: t_k = v} \hat{\tau}_{t_k, r} \indic{R_k = 0}}
\end{align*}
\EndFor
\EndFor
\State Update $\text{ELBO}(q)$ according to \eqref{eq:ELBOVEM}
\EndWhile
\Statex 2. \textbf{M-step}: Update $\hat{\boldsymbol{\pi}}_\text{VEM}$ and $\hat{\boldsymbol{\rho}}_\text{VEM}$ via
\For{$m = 1$ to $K$}
\begin{align*}
\hat{\pi}_{m} = \sum_{v \in V(n)} \hat{\tau}_{v, m}
\end{align*}
\For{$r = 1$ to $K$}
\begin{align*}
\hat{\rho}_{m, r} = \frac{\sum_{k = 1}^n \hat{\tau}_{s_k, r}\hat{\tau}_{t_k, m} \indic{R_k = 1}}{\sum_{k = 1}^n \hat{\tau}_{s_k, r}\hat{\tau}_{t_k, m} }
\end{align*}
\EndFor
\EndFor
\EndWhile
\end{algorithmic}
\end{algorithm}

\section{Model selection for an unknown number of communication types}\label{sec:unknownK}

In this section we extend the methods discussed in Section \ref{sec:knownK} to the case where the number of communication types is not known a priori. This can be viewed as a model selection problem, where the Bayesian solution places a prior on $K$ while the variationally Bayesian and EM algorithms aim to imitate marginal likelihood-based procedures.

\subsection{A prior on K}

This section extends the Bayesian solution in Section \ref{sec:BayesKknown} to making inference on the unknown number of communication classes $K$. In a fully Bayesian framework, $K$ is assigned a prior and inference is made on the posterior of $K$ given the observed data. This, however, often requires the use of complicated reversible jump MCMC (RJMCMC) algorithms to make valid posterior inference on $K$. Generically, mixture models with a prior on the number of mixture components are known as mixture of finite mixture (MFM) models. For Bayesian MFMs, \cite{miller2018mixture} derived the Dirichlet process-like properties of MFMs and proposed a collapsed Gibbs sampler that circumvented the need for RJMCMC. \cite{geng2019probabilistic} utilized a similar collapsed Gibbs sampler for learning the number of components in a stochastic block model.  Unfortunately, such collapsed Gibbs samplers require analytically marginalizing over $K$, restricting our ability to make inference on $\boldsymbol{\pi}$ without some ad-hoc post-processing of the posterior samples. Recently, a telescoping sampler has been developed by \cite{fruhwirth2021generalized} for MFMs that obviates the need to marginalize over $K$. Rather, $K$ is explicitly sampled in the scheme by distinguishing between $K$, the number of mixture components, and $K_+$, the number of \textit{filled} mixture components.

For the heterogeneous reciprocal PA model, we adopt the prior specification in \eqref{eq:priors} and additionally let $K - 1$ follow a beta-negative-binomial (BNB) distribution with parameters $c_1, c_2$ and $c_3$ as recommended by \cite{fruhwirth2021generalized}. The BNB distribution is a hierachical generalization of the Poisson, geometric, and negative-binomial distribution. If $K - 1 \sim \text{BNB}(c_1, c_2, c_3)$ then the probability mass function on $K$ is given by
\begin{align*}
p(K) = \frac{\Gamma(c_1 + K - 1)B(c_1 + c_2, K - 1 + c_3)}{\Gamma(c_1)\Gamma(K)B(c_2, c_3)}, \ K = 1, 2, \dots,
\end{align*}
where $B$ denotes the beta function. As discussed in \cite{fruhwirth2021generalized}, the BNB distribution allows for the user to specify a heavier tail on the number of mixture components which is essential in order for the telescoping sampler to mix well.
Previous analyses in \cite{geng2019probabilistic} and \cite{miller2018mixture} specify that $K - 1 \sim \text{Poisson}(1)$, which is a highly informative choice with a light tail. 

We present the telescoping sampler for heterogeneous reciprocal PA models in Algorithm \ref{alg:telescoping}. For ease of notation, we do not distinguish between $W(n)$, the communication types, and the random partition of the $|V(n)|$ nodes into $K_+$ clusters induced by $W(n)$. However, the alternating between sampling on the parameter space of the mixture distribution and the set partition space is a key aspect that allows $K$ to be directly sampled from the conditional posterior of $K$ given the partition induced by $W(n)$ (Step 3 in Algorithm \ref{alg:telescoping}). We refer to \cite{fruhwirth2021generalized} for more details on the telescoping sampler. Note that within the sampler, $K$ only decreases if one of the $K_+$ filled components loses all of its membership in Step 1. Thus, in order for the sampler to mix well, $K$ must occasionally exceed $K_+$, emphasizing the need for a heavier-tailed prior on $K$.

\cite{fruhwirth2021generalized} also present a dynamic mixture of finite mixture model where the prior on $\boldsymbol{\pi}$ is taken to be be $\text{Dirichlet}(\varphi/K, \varphi/K, \dots, \varphi/K)$ for some $\varphi > 0$. This specification would induce a sparse mixture model where a large number of mixture components $K$ would be fit, but a majority of them would be unfilled \citep{fruhwirth2019here, malsiner2016model}. In this sense, the posterior distributions on $K$ and $K_+$ would differ greatly. Though this is undesirable for learning the parameters of a mixture model, it may be useful for analyses more focused on partitioning nodes into a small number of classes with similar reciprocal behavior. 
  
\begin{algorithm}
\caption{Telescoping sampler for heterogeneous reciprocal PA with known $K$}
\label{alg:telescoping}
\begin{algorithmic}
\Require Graph $G(n)$, parameters $a$, $b$, $\eta$, $c_1$, $c_2$, $c_3$, $K$ initial/max values $K_\text{init}$, $K_\text{max}$, \# MCMC iterations $M$
\Ensure Approximate samples from the posterior $p \left( \boldsymbol{\rho}, \boldsymbol{\pi}, W(n), K \mid (e_k)_{k = 1}^n \right)$ \\
\textbf{Initialize:} Set $K = K_\text{init}$, draw $\boldsymbol{\pi}$ and $\boldsymbol{\rho}$ from \eqref{eq:priors}, draw $W_v \sim \text{Multinomial}(\boldsymbol{\pi})$ for $v \in V(n)$
\For{$i = 1$ to $M$}
\Statex 1. Sample $W(n)$ from its conditional posterior
\ForAll{$v \in V(n)$}
\State Sample $W_v$ according to
\begin{align*}
P&\left( W_v = \ell \mid \boldsymbol{\pi}, \boldsymbol{\rho}, (W_u)_{u \neq v}, (e_k)_{k = 1}^n  \right)\\
& \propto \pi_\ell \prod_{m = 1}^K  \rho_{m, \ell}^{\sum_{k:s_k = v} 1_{\{W_{t_k} = m \}}1_{\{ R_k = 1 \}}} 
(1 - \rho_{m, \ell})^{\sum_{k:s_k = v} 1_{\{W_{t_k} = m \}} 1_{\{ R_k = 0 \}}} \\
&\qquad\times\prod_{r = 1}^K \rho_{\ell, r}^{\sum_{k:t_k = v} 1_{\{W_{s_k} = r \}}1_{\{ R_k = 1 \}}} 
(1 - \rho_{\ell, r})^{\sum_{k:t_k = v} 1_{\{W_{s_k} = r \}} 1_{\{ R_k = 0 \}}},
\end{align*}
\State for $\ell = 1,\dots, K$
\EndFor
\Statex and determine the number of filled components $K_+$. Relabel the communication classes such that the first $K_+$ components are filled and the rest are empty.
\Statex 2. Sample the filled components of $\boldsymbol{\rho}$ from its conditional posterior 
\For{$m = 1$ to $K_+$}
\For{$r = 1$ to $K_+$}
\begin{align*}
\rho_{m ,r} \mid \boldsymbol{\pi}, W(n), (e_k)_{k = 1}^n  \sim \text{Beta}\Bigg(&a + \sum_{k = 1}^n 1_{\{W_{s_k = r}\}}1_{\{W_{t_k = m} \}}1_{\{R_k = 1\}},\\
& b + \sum_{k = 1}^n 1_{\{W_{s_k = r}\}}1_{\{W_{t_k = m} \}}1_{\{R_k = 0\}} \Bigg),
\end{align*}
\EndFor
\EndFor
\Statex 3. Sample $K$ from
\begin{align*}
p(K | W(n)) \propto p(K) \frac{K!}{(K - K_+)!}\frac{\Gamma(\eta K)}{\Gamma(|V(n)| + \eta K)\Gamma(\eta)^{K_+}} \prod_{r = 1 }^{K_+} \Gamma\left(\sum_{v \in V(n)} \indic{W_v = r} + \eta \right),
\end{align*}
where $K = K_+, K_+ + 1, \dots, K_\text{max}$. If $K > K_+$, generate $K - K_+$ empty components and fill the corresponding $\boldsymbol{\rho}$ components with draws from the prior $\text{Beta}(a, b)$.
\Statex 4. Sample $\boldsymbol{\pi}$ from its conditional posterior 
\begin{align*}
\boldsymbol{\pi} \mid \boldsymbol{\rho},  W(n), (e_k)_{k = 1}^n \sim \text{Dirichlet}\left( \eta + \sum_{v \in V(n)} 1_{\{W_v = 1 \}}, \dots, \eta + \sum_{v \in V(n)} 1_{\{W_v = K \}} \right)
\end{align*}
\EndFor
\end{algorithmic}
\end{algorithm}

\subsection{Imitations of the marginal likelihood}

In this section we review criteria for choosing the number of communication types $K$ for the variational methods proposed in Section \ref{sec:Variational}. A typical strategy for Bayesian model selection is choosing the model that maximizes the marginal likelihood, or the probability distribution that is obtained by integrating the likelihood over the prior distribution of the parameters. For many of the same reasons presented in Section \ref{sec:likelihood}, the marginal likelihood is not avaible for stochastic blockmodel data. Instead, for the Bayesian Variational Inference method presented in Section \ref{sec:BVI}, \cite{latouche2012variational} recommend employing the ELBO as the model selection criterion. From \eqref{eq:KL}, it can be seen that 
\begin{align*}
\text{ELBO}(q) = -\text{KL}\left(q(\cdot) \mid\mid p(\cdot | (e_k)_{k = 1}^n) \right)) + \log p\left((e_k)_{k = 1}^n \right)  \leq \log p\left((e_k)_{k = 1}^n  \right).
\end{align*}
That is, the ELBO lower bounds the marginal likelihood, and if the variational approximation to the posterior is good, the ELBO should approximate it. Though, there is no evidence that the variational approximation results in a sufficiently small KL divergence such that this is a worthwhile approximation. Regardless, this criteria is often used in practice \citep{blei2017variational}. 

For the VEM algorithm, \cite{daudin2008mixture} recommend employing the Integrated Classification Likelihood (ICL). Although the VEM algorithm is a frequentist procedure, the ICL criterion is derived by assuming a Jeffrey's prior on $\boldsymbol{\pi}$ ($\eta = 1/2$) and further employs a BIC approximation to the distribution of $(e_k)_{k = 1}^n$ given $W(n)$. The ICL for reciprocal PA models is given by
\begin{align*}
\text{ICL}(K) =& \log p( (e_k)_{k = 1}^n, \hat{W}(n) \mid \hat{\boldsymbol{\pi}}_\text{VEM}, \hat{\boldsymbol{\rho}}_\text{VEM}) - \frac{K^2}{2}\log n - \frac{K - 1}{2}\log |V(n)|,
\end{align*}
where $\hat{W}(n)$ is the modal approximation of $W(n)$ given by $\hat{W}_v = \argmax_{\ell = 1, \dots, K} \hat{\tau}_{v, \ell}$. 

\section{Simulation Studies}

In this section, we evaluate the performance of the estimation procedures presented in Sections \ref{sec:knownK} and \ref{sec:unknownK} on synthetic datasets. We evaluate the performance of estimation methods for $\boldsymbol{\pi}$ and $\boldsymbol{\rho}$ when $K$ is known, as well as the accuracy of the model selection criteria presented in Section \ref{sec:unknownK} when $K$ is unknown. When $K$ is known, we employ the Monte Carlo averages of the approximate posterior samples, the posterior means of the variational densities and the variational EM estimates as point estimators of $\boldsymbol{\pi}$ and $\boldsymbol{\rho}$ for the fully Bayesian (B), Variational Bayes (VB) and Variational EM (VEM) methods, respectively. Since the B and VB methods produce approximate posteriors, we also provide marginal coverage rates of credible intervals constructed using the element-wise 2.5\% and 97.5\% quantiles of the respective posteriors for $\boldsymbol{\pi}$ and $\boldsymbol{\rho}$. In the case of known $K$, we further provide the average Rand index for estimating $(W_v)_{v \in V(n)}$ for each method. When $K$ is unknown, we record the frequencies of the estimated $K$ under each model selection criteria. We employ the posterior mode as the estimated $K$ for the fully Bayesian method.

In each simulation, we assume non-informative priors Dirichlet$(1/2, \dots, 1/2)$ on $\boldsymbol{\pi}$ and Beta$(1/2, 1/2)$ on $\boldsymbol{\rho}$ for the VB and B methods. Although a prior is not explicitly assumed for the VEM method, the ICL model selection criterion implicitly assumes the same prior on $\boldsymbol{\pi}$, hence these choices are consistent. For the VEM algorithm, we terminate the E-step once either the ELBO has increased by less than $\epsilon = 0.01$ or the total number of iterations exceed $500$, and terminate the entire algorithm once the element-wise differences in the parameters fall below $\kappa = 0.01$. We also terminate the VB algorithm via the same conditions as in the E-step of the VEM algorithm. We run the fully Bayesian method for $M = 5{,}000$ MCMC samples, and discard the first half as burn-in. Further, when $K$ is unknown, we assume a BNB$(1, 4, 3)$ prior on $K$ as recommended by \cite{fruhwirth2021generalized} and set $K_{\text{max}} = 20$. For the variational methods, we search over $K = 1, 2, 3, 4$.

\subsection{Simulations on Pragmatic Networks}

In this section we evaluate the presented model estimation and selection criteria on networks which are generated to reflect those found in real-world applications. In particular, we generate a PA network with heterogeneous reciprocity such that $\boldsymbol{\theta} = (\alpha, \beta, \deltain, \deltaout) = (0.15, 0.8, 1, 1)$, 
\begin{align*}
\boldsymbol{\pi} = \begin{bmatrix} 0.8 \\ 0.2 \end{bmatrix} \qquad \text{and} \qquad \boldsymbol{\rho} = \begin{bmatrix} 0.5 & 0.9 \\ 0.05 & 0.2 \end{bmatrix}.
\end{align*}
This network generating process contains two groups, the first of which can be thought of as typical users, and the other can be thought of celebrities. Here typical users will often reciprocate the messages from celebrities, but a celebrity is far less likely to respond to a typical user. As one might expect, there are far more typical users than celebrities in this network. 

Table \ref{tab:Simulation3} displays the means and standard errors of the element-wise point estimators across the simulations, as well as the coverage of credible intervals produced from the B and VB methods. Here, the estimation procedures have virtually identical performance in terms of point estimation. Further, the coverage rates for the fully Bayesian method hover around the expected 95\%, while the coverage rates for the Variational Bayes method vary across the parameters. The VB method seems to have difficultly capturing the larger reciprocity $\rho_{12} = 0.9$, as well as $\pi_1 = 0.8$. The methods also perform similarly in terms of classification, as the average Rand index for the communication types are given by 0.767, 0.767 and 0.766 for the B, VB and VEM methods respectively. 

Table \ref{tab:Simulation4} displays the performance of the model selection criteria on the same preferential attachment model but for unknown $K$. For the fully Bayesian method, we initialize at $K_{\text{init}} = 4$ in order to exhibit insensitivity of the telescoping sampler to initialization. Note that the ELBO and ICL select the correct class for every simulated data set, while the fully Bayesian method has a slight tendency to over-select the number of classes. Clearly, however, analysis of such networks result in variational methods that perform comparably to the fully Bayesian method, at less computational cost.

\begin{table}
\centering
\begin{tabular}{ c c c c c c }
\hline \hline
Method & $\pi_1 = 0.8$ & $\rho_{11} = 0.5$ & $\rho_{12} = 0.9$ & $\rho_{21} = 0.05$ & $\rho_{22} = 0.2$ \\
\hline
& \multicolumn{5}{c}{Mean(SE)} \\
B & 0.803(0.003) & 0.500(0.002) & 0.900(0.004) & 0.050(0.002) & 0.198(0.010)  \\
VB & 0.805(0.003) & 0.500(0.002) & 0.896(0.004) & 0.050(0.002) & 0.198(0.010)  \\
VEM & 0.788(0.004) & 0.501(0.002) & 0.889(0.005) & 0.052(0.002) & 0.196(0.010) \\
& \multicolumn{5}{c}{\% Coverage} \\
B & 98 & 93 & 94 & 92 & 94 \\
VB & 50 & 90 & 70 & 87 & 92 \\
\hline
\end{tabular}
\caption{Average point estimates and standard errors for 100 networks generated from a PA model with $\boldsymbol{\theta} = (0.15, 0.8, 1, 1)$. Coverage rates for equal-tailed credible intervals produced by the B and VB methods are also provided.}
\label{tab:Simulation3}
\end{table}

\begin{table}
\centering
\begin{tabular}{ c c c c c  }
\hline \hline
Method & \multicolumn{4}{c}{$\hat{K}$} \\
\hline
 & 1 & 2 & 3 & 4 \\
B & 0 & 68 & 31 & 1 \\
VB & 0 & 100 & 0 & 0 \\
VEM & 0 & 100 & 0 & 0 \\
\hline
\end{tabular}
\caption{Estimated $K$ from 100 networks generated from a PA model with $\boldsymbol{\theta} = (0.15, 0.8, 1, 1)$ and $\boldsymbol{\pi}$ and $\boldsymbol{\rho}$ as in Table \ref{tab:Simulation3}.}
\label{tab:Simulation4}
\end{table}

We continue our simulations by evaluating the performance of the estimation procedures on 100 synthetic networks generating from a PA network with heterogeneous reciprocity such that $\boldsymbol{\theta} = (0.15, 0.8, 1, 1)$ but now
\begin{align*}
\boldsymbol{\pi} = \begin{bmatrix} 0.8 \\ 0.2 \end{bmatrix} \qquad \text{and} \qquad \boldsymbol{\rho} = \begin{bmatrix} 0.5 & 0 \\ 0.05 & 0.2 \end{bmatrix}.
\end{align*}
Note that the only difference from this simulation set-up and the previous one is that $\rho_{12}$ has decreased from $0.9$ to $0$. The inclusion of $0$ into the $\boldsymbol{\rho}$ matrix is motivated by the data example in Section~\ref{sec:Facebook}, where we find a group of users that do not receive reciprocal edges. This set-up is analogous to a diagonally-dominant stochastic block model where users are likely to communicate within groups but not across groups. 

Table \ref{tab:Simulation5} displays the point estimates for all three methods, along with the coverage probabilities for the B and VB methods. With the decrease in $\rho_{12}$, the variational methods struggle to recover $\rho_{22}$. This is sensible since class $2$ communicating with class $2$ should be the least common communication type according to $\boldsymbol{\pi}$ and, unlike the case when $\rho_{12} = 0.9$, the difference between the communication classes is not obvious. Otherwise, the estimation accuracy of the other parameters is relatively consistent across all the methods. Although coverage rates are similar to Table~\ref{tab:Simulation3}, we also observe a reduction in the coverage of $\rho_{22}$. Evidently, equal-tailed credible intervals are a poor choice for capturing $\rho_{12}$ and if one had prior knowledge on the behavior of $\boldsymbol{\rho}$, a highest posterior density interval would be a sensible choice. The average Rand index for the B, VB and VEM methods are given by 0.763, 0.763 and 0.760, respectively, again indicating that the methods classify similarly when the number of edges far exceeds the number of nodes. 

Table \ref{tab:Simulation6} displays the performance of the model selection criteria presented in Section \ref{sec:unknownK} for the preferential attachment model as in \ref{tab:Simulation5}. Again, $K$ is initialized at $K_\text{init} = 4$ for the fully Bayesian method. Note that, again, the variational methods select the correct number of clusters in each simulation, while the telescoping sampler has a slight tendency to overfit. 

\begin{table}
\centering
\begin{tabular}{ c c c c c c }
\hline \hline
Method & $\pi_1 = 0.8$ & $\rho_{11} = 0.5$ & $\rho_{12} = 0.00$ & $\rho_{21} = 0.05$ & $\rho_{22} = 0.2$ \\
\hline
& \multicolumn{5}{c}{Mean(SE)} \\
B & 0.802(0.003) & 0.500(0.002) & 0.001(0.001) & 0.051(0.003) & 0.201(0.019)  \\
VB & 0.805(0.003) & 0.501(0.002) & 0.001(0.001) & 0.051(0.003) & 0.174(0.019)  \\
VEM & 0.791(0.010) & 0.503(0.003) & 0.006(0.002) & 0.054(0.004) & 0.154(0.020) \\
& \multicolumn{5}{c}{\% Coverage} \\
B & 98 & 90 & 0 & 98 & 91 \\
VB & 49 & 88 & 0 & 86 & 47 \\
\hline
\end{tabular}
\caption{Average point estimates and standard errors for 100 networks generated from a PA model with $\boldsymbol{\theta} = (0.15, 0.8, 1, 1)$. Coverage rates for equal-tailed credible intervals produced by the B and VB methods are also provided.}
\label{tab:Simulation5}
\end{table}

\begin{table}
\centering
\begin{tabular}{ c c c c c }
\hline \hline
Method & \multicolumn{4}{c}{$\hat{K}$} \\
\hline
 & 1 & 2 & 3 & 4 \\
B & 0 & 78 & 20 & 2 \\
VB & 0 & 100 & 0 & 0 \\
VEM & 0 & 100 & 0 & 0 \\
\hline
\end{tabular}
\caption{Estimated $K$ from 100 networks generated from a PA model with $\boldsymbol{\theta} = (0.15, 0.8, 1, 1)$ and $\boldsymbol{\pi}$ and $\boldsymbol{\rho}$ as in Table \ref{tab:Simulation6}.}
\label{tab:Simulation6}
\end{table}

\subsection{Comparisons to the SBM}

In this section we evaluate the same estimation and model selection procedures on synthetic networks with a comparably low number of edges relative to the number of nodes. Such networks serve to highlight the additional difficulties faced by estimating reciprocal PA models compared to stochastic block models. We simulate 100 preferential attachment networks of size $n = 30{,}000$ from a PA model with $\boldsymbol{\theta} = (\alpha, \beta, \deltain, \deltaout) = (0.75, 0, 0.8, 0.8)$ and the reciprocal component governed by
\begin{align*}
\boldsymbol{\pi} = \begin{bmatrix} 0.6 \\ 0.4 \end{bmatrix} \qquad \text{and} \qquad \boldsymbol{\rho} = \begin{bmatrix} 0.1 & 0.4 \\ 0.5 & 0.8 \end{bmatrix}.
\end{align*}
\cite{wang2022random} have showed that, under suitable conditions, such heterogeneous reciprocal PA models with $\beta = 0$ exhibit networks with out/in-degrees that exhibit a complex extremal dependence structure (see Appendix A for more details). Additionally, since $\beta = 0$, such models allow for the complete observation of the reciprocal edge events as there are no $J_k = 2$ edges that could be mistaken as reciprocal edges.

Here we assume $K$ is known. Table \ref{tab:Simulation1} displays the average value of the point estimates of $\boldsymbol{\pi}$ and $\boldsymbol{\rho}$ for each method, as well as their associated standard errors. Clearly, the fully Bayesian method outperforms both the VB and VEM methods by producing accurate point estimates with lower standard errors. Additionally, the coverage rates for the fully Bayesian method are near the expected 95\% level, while the VB method produces posteriors which do not reliably capture the true $\boldsymbol{\pi}$ and $\boldsymbol{\rho}$. The fully Bayesian method also dominates in terms of classification, as the average Rand index for the communication types are given by 0.590, 0.583 and 0.552 for the B, VB and VEM methods, respectively. 

The superiority of the fully Bayesian method compared to the variational methods is unsurprising in this setting. Although variational methods exhibit strong point estimation for stochastic block models, estimation for PA models with heterogeneous reciprocity is an inherently harder problem. Namely, in a directed stochastic block model, each node has the opportunity to connect to every other node in the network. This results in $m(m-1)$ many potential edges for $m$ many nodes in the network. For the PA model, one expects the number of potential edges to scale linearly with the number of nodes. Thus, there is inherently less observed information that can be leveraged to learn the latent communication classes. Such lack of information induces a multimodal ELBO, and therefore the variational methods struggle to find a global optimum. The fully Bayesian method is better able to incorporate this uncertainty since it is sampling from, not optimizing, a multimodal posterior.

Table \ref{tab:Simulation2} displays the performance of the model selection criteria for 100 networks generated under the same PA model. For the fully Bayesian method, we initialize at $K_\text{init} = 1$. Despite the poor performance of the VB method at the parameter estimation, it captures the true $K = 2$ the most often, indicating that the ELBO is a good model selection criteria. The VEM algorithm always chooses $K = 1$, though we expect that this is again due to the lack of information in the data. The likelihood associated with $\boldsymbol{\pi}$ has a much larger role in the ICL for PA models than in stochastic block models. This, combined with the poor estimation of the classes for known $K$, results in the poor performance of the ICL criteria.  

\begin{table}
\centering
\begin{tabular}{ c c c c c c }
\hline \hline
Method & $\pi_1 = 0.6$ & $\rho_{11} = 0.1$ & $\rho_{12} = 0.5$ & $\rho_{21} = 0.4$ & $\rho_{22} = 0.8$ \\
\hline
& \multicolumn{5}{c}{Mean(SE)} \\
B & 0.604(0.015) & 0.102(0.011) & 0.500(0.018) & 0.400(0.016) & 0.800(0.020)  \\
VB & 0.587(0.028) & 0.168(0.058) & 0.523(0.035) & 0.331(0.0216) & 0.718(0.082)  \\
VEM & 0.599(0.073) & 0.121(0.005) & 0.441(0.091) & 0.286(0.077) & 0.686(0.074) \\
& \multicolumn{5}{c}{\% Coverage} \\
B & 95 & 93 & 98 & 93 & 92 \\
VB & 19 & 0 & 6 & 11 & 0 \\
\hline
\end{tabular}
\caption{Average point estimates and standard errors for 100 networks generated from a PA model with $\boldsymbol{\theta} = (0.75, 0, 0.8, 0.8)$. Coverage rates for equal-tailed credible intervals produced by the B and VB methods are also provided.}
\label{tab:Simulation1}
\end{table}

\begin{table}
\centering
\begin{tabular}{ c c c c c c }
\hline \hline
Method & \multicolumn{5}{c}{$\hat{K}$} \\
\hline
 & 1 & 2 & 3 & 4 & 5  \\
B & 0 & 83 & 12 & 4 & 1 \\
VB & 0 & 95 & 5 & 0 & 0 \\
VEM & 100 & 0 & 0 & 0 & 0 \\
\hline
\end{tabular}
\caption{Estimated $K$ from 100 networks generated from a PA model with $\boldsymbol{\theta} = (0.75, 0, 0.8, 0.8)$ and $\boldsymbol{\pi}$ and $\boldsymbol{\rho}$ as in Table \ref{tab:Simulation1}.}
\label{tab:Simulation2}
\end{table}

\section{Data Example}\label{sec:Facebook}

\begin{figure}
\centering
\includegraphics[scale=0.65]{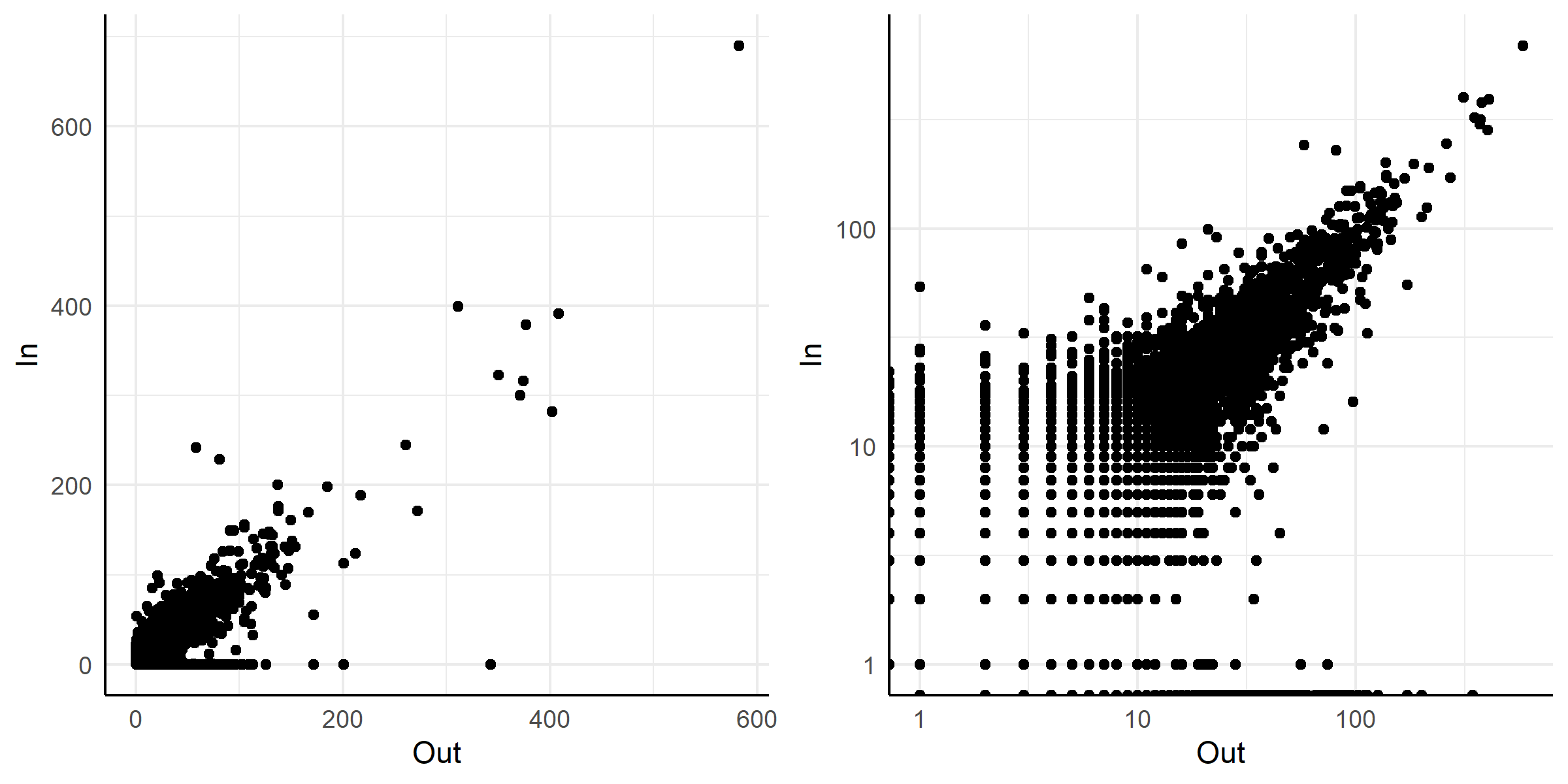}
\caption{Out/in-degree plot for the Facebook wallpost data}
\label{fig:outin}
\end{figure}

\begin{figure}
\centering
\includegraphics[scale=0.65]{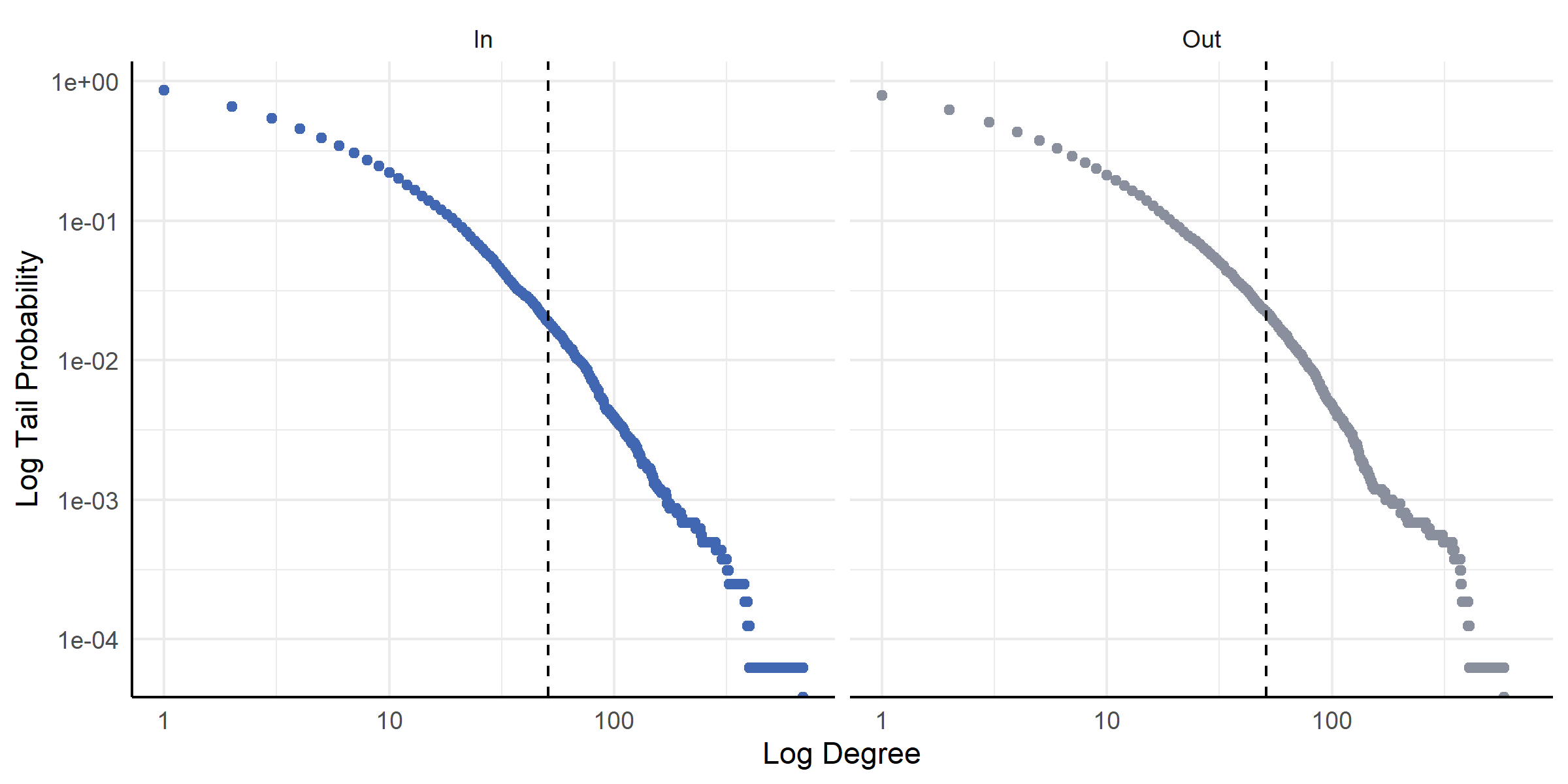}
\caption{Plot of empirical tail probability function for the Facebook wallpost degrees on a log base 10 scale}
\label{fig:log_tail}
\end{figure}

Now we apply the heterogeneous reciprocal PA model to the Facebook wall post data from KONECT analyzed in \cite{viswanath2009evolution} and \cite{cirkovic2022preferential}. The Facebook wall post data tracks a group of users in New Orleans and their wallposts from September 9th, 2004 to January 22nd, 2009. The network is temporal: when user $u$ posts to user $v$'s wall, a directed edge $(u,v)$ is generated and the timestamp of the post is recorded. The full dataset consists of 876,933 wallposts and 46,952 users. In Figure \ref{fig:outin}, we display the out/in-degree of each user in a trimmed version of the network; we postpone the discussion of the data cleaning procedure to the following paragraph. Note that upon first observation, the degree distribution indicates the existence of two populations that exhibit differing reciprocal behavior. The first group, concentrated on the out-degree axis, mostly post on other users' walls while not receiving any posts on their own. The second group both sends and receives wall posts at a commensurate rate. Further, the marginal out/in-degree distributions exhibit power law tails as indicated by Figure \ref{fig:log_tail} where, on the log-log scale, the empirical tail functions seems to scale linearly with large degrees. 

In \cite{cirkovic2022preferential}, the Facebook wall post data was analyzed assuming that each user exhibited homogeneous reciprocal behavior. In the wording of Section \ref{sec:PAmodel}, it was assumed that $\boldsymbol{\pi} \equiv 1$ and $\boldsymbol{\rho} \equiv \rho \in \mathbb{R}$. In doing so, the users concentrated on the out-degree axis in Figure \ref{fig:outin} were excluded from the analysis as the homogeneous model could not model the observed hetereogenous reciprocal behavior. Additionally, by virtue of extreme value-based methods being sensitive to the choice of seed graph, \cite{cirkovic2022preferential} also removed nodes that became inactive as the graph evolved, a phenomena not modeled by the proposed PA model. The likelihood-based methodology in \cite{cirkovic2022preferential} returned a homogeneous reciprocity estimate of $\hat{\rho} = 0.28$. The flexibility provided by the heterogeneous reciprocal PA model aims to capture the additional, intricate dynamics underlying the Facebook wall post data not previously considered in \cite{cirkovic2022preferential}. 

According the analysis of the Facebook wallpost data in \cite{viswanath2009evolution}, there observes a sudden uptick in the number of wall posts from July 2008 and onwards. They conjecture that this uptick is likely due to a Facebook redesign, introduced in July, that allowed users to interact with more wall posts through friend feeds. This likely results in a distributional shift in the network's evolution, and thus we discard the portion of the network observed beyond June, 2008, resulting in a network with 22,286 nodes 165,776 edges. This observation, however, may lead to additional analyses via changepoint detection \citep[see][for example]{banerjee2018fluctuation, bhamidi2018change, cirkovic2022likelihood}. Additionally, the evolution of the PA network specified in Section \ref{sec:PAmodel} posits that every new edge must attach to at least one node that was previously observed in the network evolution. In order to better adhere to this assumption we define a sequence of networks by first selecting the node with the largest total degree and pairing it with the first node it makes a connection with to create a seed graph $G(0)$. Then, we only retain the edges $(u, v)$ that are (i) observed after the introduction of the seed graph and (ii) $u \in V(k - 1)$ or $v \in V(k - 1)$. This trimming procedure results in a connected network of 16,099 nodes and 123,920 edges that could have realistically been generated by a heterogeneous reciprocal PA model.

The reciprocal PA model assumes that reciprocal edges $(t_k, s_k)$ are generated instantaneously with their parent edge $(s_k, t_k)$. However in the Facebook wall post network, it is likely that in the time between reciprocated wall posts, wall posts between other users have been generated. Thus, similar to \cite{cirkovic2022preferential}, we employ window estimators to identify reciprocal edges. That is, if $e_k = (s_k, t_k)$ has a reciprocal counterpart $(t_k, s_k)$ appear in 24 hours, we attribute the event $R_k = 1$ to the edge $e_k$, redefine $e_k := e_k \cup (t_k, s_k)$ and drop $(t_k, s_k)$ from the edgelist. This results in an edgelist that is in alignment with Section \ref{sec:PAmodel}. 

To conclude our exploratory data analysis, we study the tail behavior of the out/in-degrees for the trimmed Facebook network. We employ the minimum distance procedure \citep{clauset2009power} on the total degrees to obtain a threshold beyond which a power-law tail for the in/out-degree can be safely assumed. The minimum distance procedure computes a tail threshold of $51$. Note that computing the tail threshold on the total degree implicitly assumes that the out/in-degree tails have the same power-law index. We find this to be a reasonable assumption as indicated by the similarity of the empirical tail functions in Figure \ref{fig:log_tail}. In fact, using a threshold of $51$, the tail index estimates for the out/in-degrees are $2.212$ and $2.231$, respectively. Further observation of Figure \ref{fig:outin} indicates that, beyond this threshold, there is an extremal dependence structure in the out/in-degree distribution; nodes with total degree larger than $51$ tend to cluster  around multiple lines through the origin. This extremal dependence structure is further analyzed in Appendix A.

We fit the VEM, VB and fully Bayesian methods to the Facebook wall post network. For the VEM algorithm, we terminate the variational E-step when the increase in the ELBO is less than $\epsilon = 0.1$ and terminate the overall algorithm once the largest absolute difference in the estimated components of $\boldsymbol{\pi}$ and $\boldsymbol{\rho}$ between M-steps falls below $\kappa = 0.001$. For the Bayesian methods, we again assume non-informative priors on $\boldsymbol{\pi}$ and $\boldsymbol{\rho}$. Analogous to the VEM algorithm, we terminate the VB procedure once the change in the ELBO falls below $\epsilon = 0.1$. Both the VEM and VB methods are fit for $K = 1, \dots, 10$. The telescoping sampler for the fully Bayesian method is ran for $M = 100{,}000$ MCMC iterates, where the first $90{,}000$ iterates are discarded as burn-in. Within the telescoping sampler, we set $K_\text{max} = 20$. 

\begin{figure}
\centering
\includegraphics[scale=0.6]{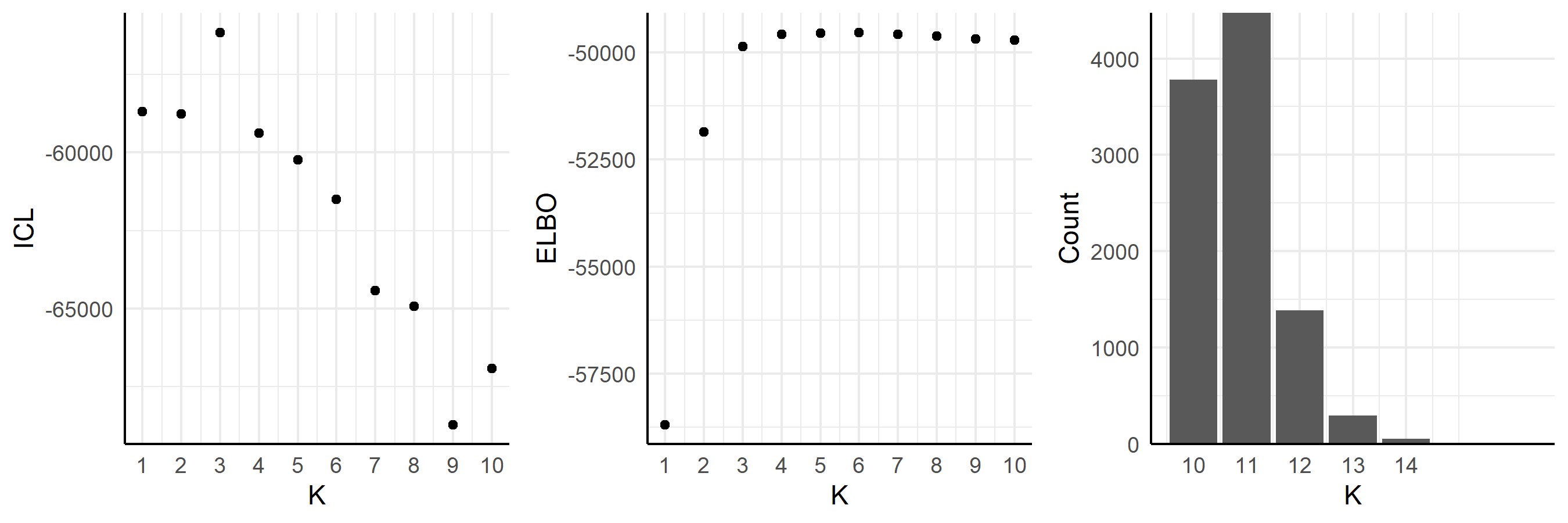}
\caption{ICL, ELBO and posterior on K from the VEM, VB and fully Bayesian methods. For the VEM and VB algorithms, we consider $K = 1, \dots, 10$.}
\label{fig:K_plots}
\end{figure}

The global PA parameters $\boldsymbol{\theta}$ are estimated by maximizng the likelihood $p(\cdot \mid \boldsymbol{\theta})$. Maximum likelihood returns $(\hat{\alpha}, \hat{\beta}, \hat{\delta}_\text{in}, \hat{\delta}_\text{out}) = (0.071, 0.829, 1.756, 1.571)$. The small values of $\hat{\delta}_\text{in}$ and $\hat{\delta}_\text{out}$ indicate that preferential attachment is indeed a viable mechanism to describe how users send and receive wall posts. Analyzing the reciprocal component of the model, the VEM algorithm identifies 3 clusters, while the VB and fully Bayesian algorithm identify 6 and 11 clusters, respectively. Figure \ref{fig:K_plots} displays the ICL, ELBO and posterior of $K$ for the VEM, VB and fully Bayesian methods. The ICL criterion clearly identifies $K = 3$ as the choice that optimally balances model parsimony with fidelity to the data. Though the VB method chooses $K = 6$, we note that the ELBO for the VB method becomes very flat at $K = 4$, indicating that perhaps a simpler model may fit the data nearly as well as the model with $K = 6$ mixture components. We suspect that the fully Bayesian method overfits the number of mixture components due to model misspecification. It is unlikely that the Facebook wallpost data exactly follows the specification in Section \ref{sec:PAmodel}. For example, there is empirical evidence that the degree of each node may influence reciprocal behavior \citep{cheng2011predicting}. There is strong evidence that mixtures of finite mixtures do not reliably learn the number of mixture components under model misspecification \citep{cai2021finite, miller2018robust}. 

\begin{figure}
\centering
\includegraphics[scale=0.4]{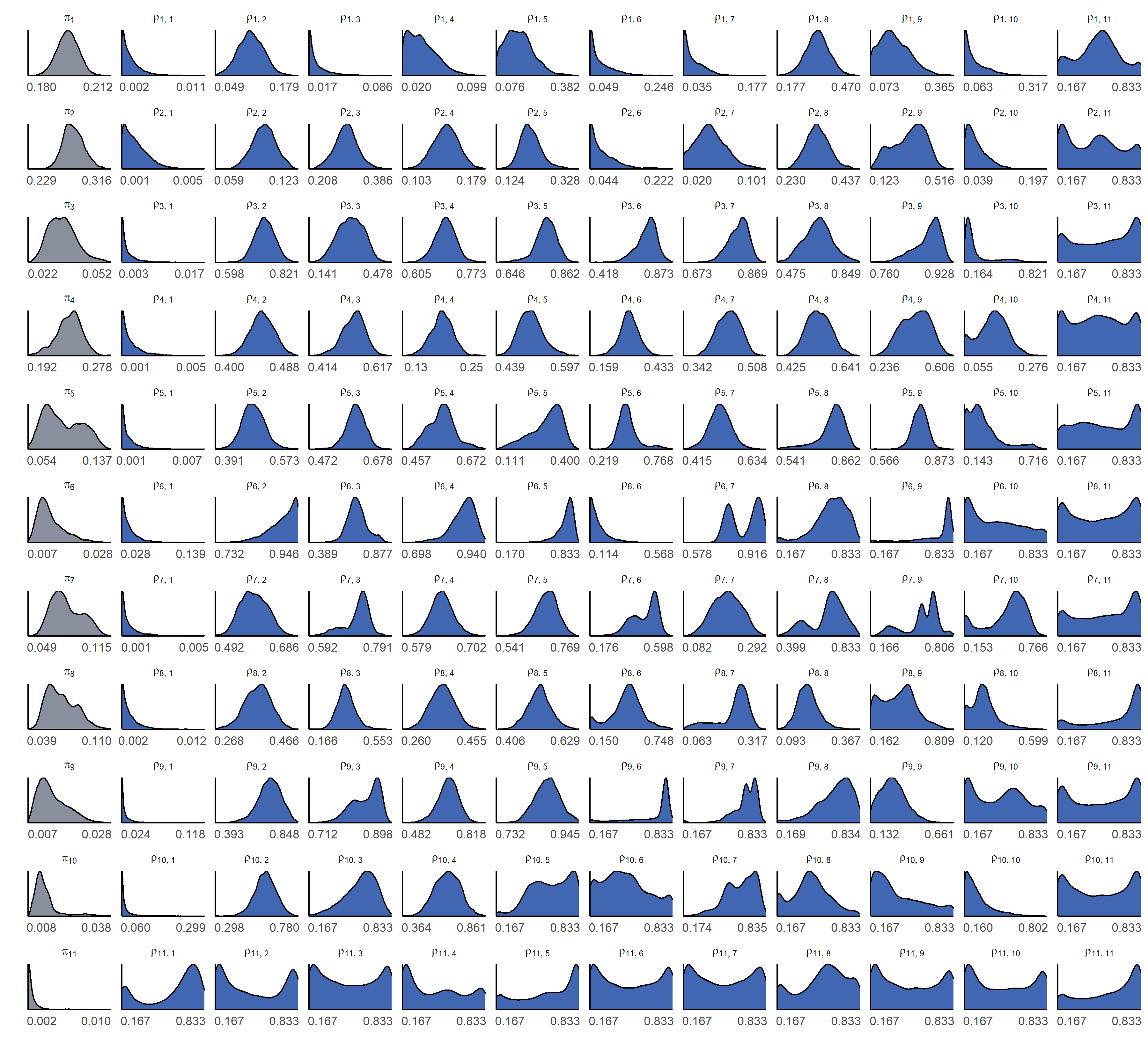}
\caption{Marginal posteriors for $\boldsymbol{\pi}$ and $\boldsymbol{\rho}$ using the telescoping sampler.}
\label{fig:marginal_posterior}
\end{figure}

The estimates of $\boldsymbol{\pi}$ and $\boldsymbol{\rho}$ for VEM and VB are 
\begin{equation}\label{eq:est}
\begin{split}
\hat{\boldsymbol{\pi}}_{\text{VEM}} = 
\begin{bmatrix}
0.538 \\ 0.251 \\ 0.211 
\end{bmatrix},
\qquad 
&\hat{\boldsymbol{\rho}}_{\text{VEM}} = 
\begin{bmatrix}
0.242 & 0.273 & 0.001 \\
0.597 & 0.650 & 0.001 \\
0.0701 & 0.053 & 0.001
\end{bmatrix} \\
\hat{\boldsymbol{\pi}}_{\text{VB}} = 
\begin{bmatrix}
0.122 \\
0.285 \\
0.153 \\
0.060 \\
0.197 \\
0.184 
\end{bmatrix},
\qquad 
&\hat{\boldsymbol{\rho}}_{\text{VB}} = 
\begin{bmatrix}
0.088 & 0.094 & 0.084 & 0.038 & 0.001 & 0.083 \\
0.383 & 0.427 & 0.431 & 0.182 & 0.001 & 0.375 \\
0.670 & 0.699 & 0.718 & 0.433 & 0.001 & 0.641 \\
0.467 & 0.464 & 0.499 & 0.214 & 0.002 & 0.437 \\
0.089 & 0.089 & 0.059 & 0.036 & 0.005 & 0.082 \\
0.206 & 0.225 & 0.230 & 0.098 & 0.001 & 0.201
\end{bmatrix}
\end{split}
\end{equation}
We also plot the marginal posteriors for $\boldsymbol{\pi}$ and $\boldsymbol{\rho}$ obtained by the telescoping sampler in Figure \ref{fig:marginal_posterior}. Note that all three methods identify a group of nodes that receive nearly no reciprocal edges as indicated by a column of near-zero estimates in $\boldsymbol{\rho}$. Additionally, the telescoping sampler seems to overfit the number of clusters by producing a cluster whose mixture weight, $\pi_{11}$, has a posterior mean of $0.0008$. Class 11 also has marginal posteriors for $\boldsymbol{\rho}$ that clearly have not yet mixed well. Hence, we caution making inference on node classes that either have a small number of nodes in them, or continually drop in and out of the sampler.

\begin{figure}
\centering
\includegraphics[scale=0.5]{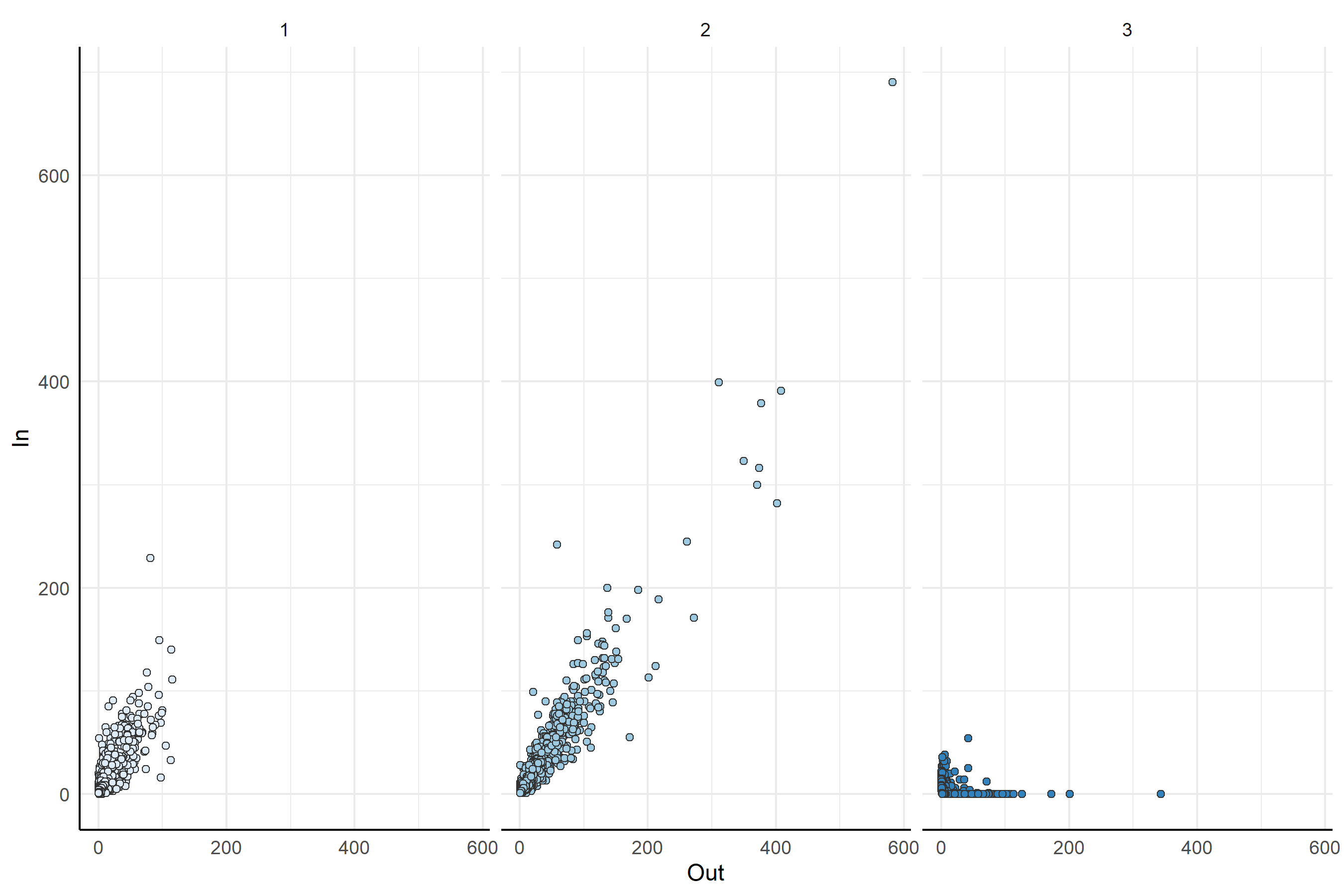}
\caption{Reciprocal components identified by the VEM algorithm}
\label{fig:clustering}
\end{figure}

Figure \ref{fig:clustering} displays the degree distribution of the trimmed Facebook wall post network, grouped by the VEM cluster estimates. The VEM algorithm clearly identifies cluster 3 as nodes that do not receive reciprocal edges. Despite, cluster 2 having a heavier tail, and clusters 1 and 2 tend to concentrate in similar regions of $\mathbb{R}_+$. Further, the similarity of the estimates $\hat{\boldsymbol{\rho}}_\text{VEM}$ indicate that classes 1 and 2 engage in similar reciprocal behavior. 

\begin{figure}
\centering
\includegraphics[scale=0.5]{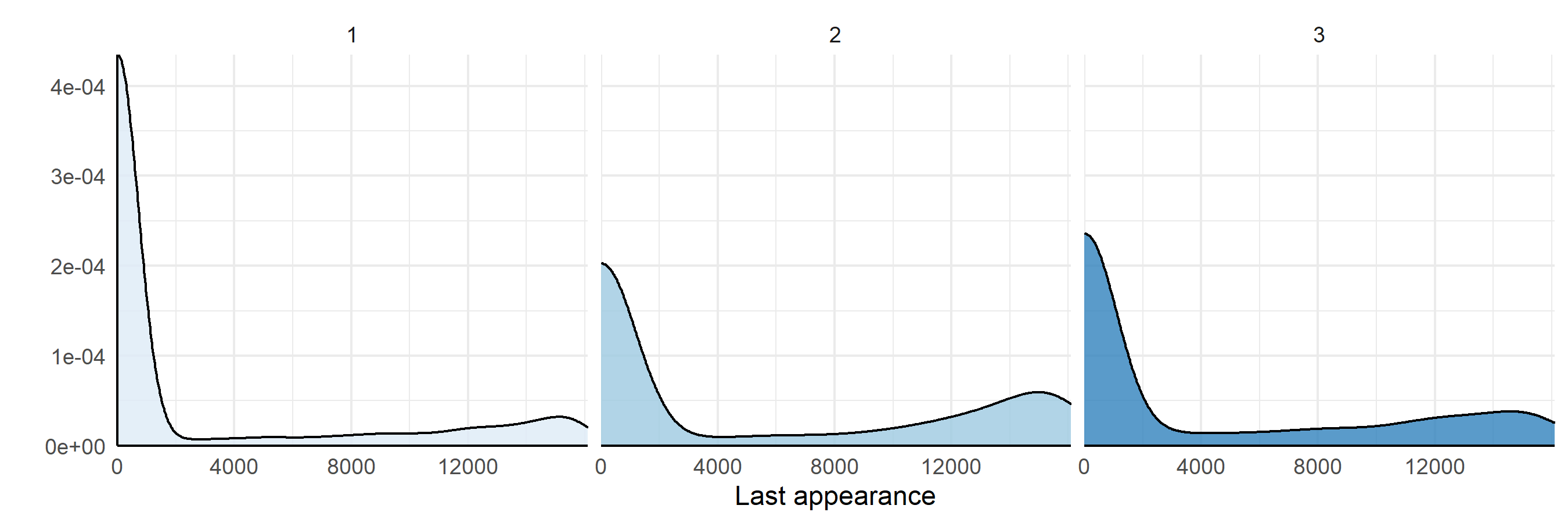}
\caption{Density plots for last appearance time (by class) of each node that posted more than once in the network}
\label{fig:last_appear}
\end{figure}

These visual measures warrant further inspection on the differences between class 1 and 2. Figure \ref{fig:last_appear} displays the discrete time of the last post made by each node in the network that posts more than once. Note that nodes in class 1 are more likely to become inactive in the early period of the network evolution. These inactive nodes were noted by \cite{cirkovic2022preferential} and \cite{viswanath2009evolution} as well. The lighter tails of class 1 thus can be explained by the relatively short lifetimes of the nodes, as such nodes do not have as long enough time to send and receive wallposts. The VEM algorithm may have picked up on this inactivity by proxy. Though, such observations warrant extension to a preferential attachment model that incorporates nodes that become inactive over time.

\section{Conclusion}\label{sec:conclusion}

In this paper, we outline a preferential attachment model with heterogeneous reciprocity, and offer three methods for fitting the model to both simulated and real-world networks. Through simulations, we find that when analyzing networks that have many edges compared to the number of nodes, the variational alternatives offer similar performance to the fully Bayesian method in terms of point estimation at less computational cost. However, the credible intervals generated by the fully Bayesian method more reliably capture the true data-generating parameters. We also compare the ability of each method to select the number of communication classes in heterogeneous reciprocal PA networks. Generally speaking, when the number of edges are again large compared to the number of nodes, all three methods consistently choose the true number of classes, with the fully Bayesian method having a slight tendency to overfit. We then showcase the ability of the heterogeneous reciprocal PA model to capture non-uniform reciprocal behavior across users in the Facebook wallpost network. The proposed model clearly offers the additional flexibility needed to model such data.

Upon analyzing the Facebook wallpost network, we find that the VEM algorithm uncovered two reciprocal classes that engage in somewhat similar reciprocal behavior, though one of the classes consisted of more inactive users. The propensity of some users to become inactive in a network as it evolves over time is a common feature of many networks, and warrants the extension of the preferential attachment model to account for such behavior. In future work, we will also consider models that allow for users to become inactive as the network grows over time.


\acks{The research work was partly supported by NSF Grant DMS-2210735.}


\newpage

\appendix
\section*{Appendix A: Statistical Tools for Multivariate Extremes}\label{appA}
Here we detail, in a non-technical fashion, some tools used to analyze data subject to extremal observations. For more rigorous treatments, we refer to the works of \cite{beirlant2004statistics} and \cite{resnick2007heavy}. A central goal in the study of multivariate extremes is to identify how extremes cluster. In other words, if one or more components of a random vector is large, how likely is that the other components of the random vector will also be large? For PA models with homogeneous reciprocity, \cite{cirkovic2022preferential} proved that the extremal out/in-degrees tend to cluster on a line through the origin. With heterogeneous reciprocity, \cite{wang2022random} proved that the model with $\beta = 0$ generates extreme out/in-degrees that concentrate on multiple lines through the origin. 

\begin{figure}
\centering
\includegraphics[scale=0.65]{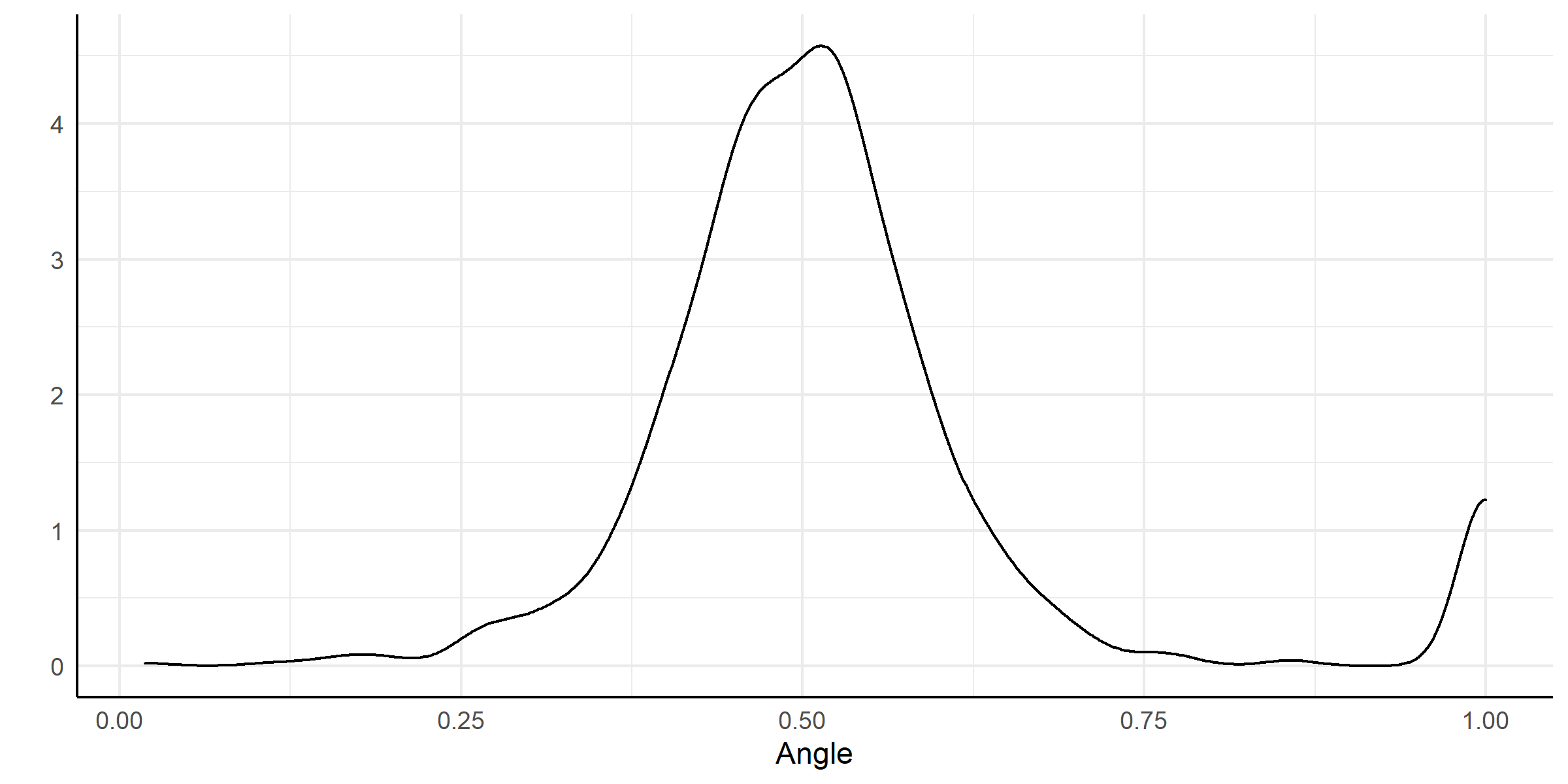}
\caption{Angular density for the Facebook wallpost data with threshold chosen via the minimum distance method.}
\label{fig:angular_density}
\end{figure}

An exploratory tool used to identify where such extremes cluster in $\mathbb{R}^2_+$ is the \textit{angular density}, a plot of the angles 
$$\Theta_r \equiv \left\lbrace D^\text{out}_v(n)/(D^\text{out}_v(n) + D^\text{in}_v(n)) : v \in V(n), D^\text{out}_v(n) + D^\text{in}_v(n) > r \right\rbrace$$
 for some large threshold $r$. Intuitively, if the angular density concentrates mass around some point in $(0, 1)$, then one would expect extremes to cluster on a line through the origin. On the other hand, if the angular density only places mass on the set $\{0, 1\}$, then the out/in-degrees are \textit{asymptotically independent}; a large in-degree does not necessarily imply a large out-degree, and vice versa. Figure \ref{fig:angular_density} displays the angular density for the Facebook wallpost data analyzed in Section \ref{sec:Facebook}.

When the angular density concentrates mass on the set $\{0.5, 1 \}$, it indicates the existence of two extremal populations: one that has approximately equal in/out-degree, and another that has high out-degree but small in-degree. The threshold for $\Theta_r$ was chosen as $r = 51$ by the minimum distance method applied to the total degrees \citep{clauset2009power}. The minimum distance method chooses a threshold that minimizes the Kolmogorov-Smirnov distance between a power-law tail and the emprical tail of the observations above the threshold. Note that the angular density is naturally sensitive to the choice of $r$. If $r$ is chosen too large, some extremal features of the data may be passed over, while if $r$ is chosen too small, the extremal behavior will be corrupted by non-extremal observations. 

We now have the tools to describe the intialization of the VEM algorithm for a fixed $K$ presented in Section \ref{sec:VEM}. First, the set $\Theta_r$ is constructed via threshold $r$ chosen by the minimum distance method available in R package \texttt{igraph} \citep{csardi2006igraph}. We then employ $K$-means on the set $\Theta_r$ to determine an initial clustering of nodes. Note that this only clusters nodes with total degree larger than $r$. This clustering is then used to compute empirical class probabilities $(\hat{\pi}_r)_{r = 1}^K$ and empirical reciprocities $(\hat{\rho}_{m, r})_{m, r = 1}^K$. Note that $(\hat{\rho}_{m, r})_{m, r = 1}^K$ is computed only on edges that connect nodes which both have total degree larger than $r$. $(\hat{\pi}_r)_{r = 1}^K$ and $(\hat{\rho}_{m, r})_{m, r = 1}^K$ are thus used as initial parameter values and the the initial $(\tau_{w, \ell})_{w \in V(n), \ell \in \{1, \dots, K \}}$ are chosen according to a uniform distribution on the $K$-simplex. The full initialization algorithm is given in Algorithm \ref{alg:VEMinit}.

\begin{algorithm}
\caption{Initialization of VEM for heterogeneous reciprocal PA}
\label{alg:VEMinit}
\begin{algorithmic}
\Require Graph $G(n)$, \# communication types $K$
\Ensure Initial variational EM estimates $\hat{\boldsymbol{\pi}}_\text{VEM}$ and $\hat{\boldsymbol{\rho}}_\text{VEM}$ 
\State 1. Compute the tail threshold $r$ according to the minimum distance procedure.
\State 2. Construct the sets 
\begin{align*}
\aleph_r &= \left\lbrace v \in V(n) : D^\text{out}_v(n) + D^\text{in}_v(n) > r  \right\rbrace \\
\Theta_r &= \left\lbrace D^\text{out}_v(n)/(D^\text{out}_v(n) + D^\text{in}_v(n)) : v \in \aleph_r \right\rbrace
\end{align*}
\State 3. Employ $K$-means on $\Theta_r$ to form initial communication class estimates $\hat{W}_v$ for $v \in \aleph_r$
\State 4. Form initial VEM estimates via
\For{$m = 1$ to $K$}
\begin{align*}
\hat{\pi}_{m} = \frac{1}{|\aleph_r|} \sum_{v \in \aleph_r} 1_{\{\hat{W}_v = m \}}
\end{align*}
\For{$r = 1$ to $K$}
\begin{align*}
\hat{\rho}_{m, r} = \frac{\sum_{k : \hat{W}_{s_k} = r, \hat{W}_{t_k} = m } \indic{R_k = 1}}{\left| \left\lbrace k : \hat{W}_{s_k} = r, \hat{W}_{t_k} = m \right\rbrace \right| }
\end{align*}
\EndFor
\EndFor
\end{algorithmic}
\end{algorithm}

\section*{Appendix B: Sample Derivations for the VEM Algorithm}\label{appB}

In this appendix we present some sample derivations for the variational EM algorithm presented in Section \ref{sec:VEM}. We note that the derivations are very similar to those of \cite{daudin2008mixture} and \cite{latouche2012variational}, though we reformulate them in our setting for convenience. The same type of calculations can be employed to derive the variational Bayes algorithm.

\subsection*{Derivation of the ELBO}

In this section we derive \eqref{eq:ELBOVEM}. Recall that we posit the mean-field variational family on the communication types $W(n)$ given by
\begin{align*}
q(W(n)) = \prod_{v \in V(n)} q_v(W_v).
\end{align*} 
Then, the ELBO is given by
\begin{align*}
\text{ELBO}(q, \boldsymbol{\pi}, \boldsymbol{\rho}) = E_q \left[ \log p\left( W(n), (e_k)_{k = 1}^n \mid \boldsymbol{\pi}, \boldsymbol{\rho} \right) \right] - E_q\left[\log q(W(n))\right]. 
\end{align*}
Focusing on the first term, recall that the log-likelihood is given by
\begin{align*}
&\log p\left( W(n), (e_k)_{k = 1}^n \mid \boldsymbol{\pi}, \boldsymbol{\rho} \right)\\
 =& \sum_{k = 1}^n \sum_{r = 1}^K \left( 1_{\{ J_k = 1 \}} 1_{\{W_{s_k} = r \}} + 1_{\{ J_k = 3 \}} 1_{\{W_{t_k} = r \}} \right) \\
&+ \sum_{k = 1}^n \sum_{r = 1}^K \sum_{m = 1}^K 1_{\{ R_k = 1 \}} 1_{\{W_{s_k} = r \}} 1_{\{W_{t_k} = m \}} \log \rho_{m, r} \\
&+ \sum_{k = 1}^n \sum_{r = 1}^K \sum_{m = 1}^K 1_{\{ R_k = 0 \}} 1_{\{W_{s_k} = r \}} 1_{\{W_{t_k} = m \}} \log (1 - \rho_{m, r}).
\end{align*}
Taking an expectation with respect to $q$ gives that
\begin{align*}
E_q & \left[ \log p\left( W(n), (e_k)_{k = 1}^n \mid \boldsymbol{\pi}, \boldsymbol{\rho} \right) \right]\\
 =& \sum_{k = 1}^n \sum_{r = 1}^K \left( 1_{\{ J_k = 1 \}} E_q \left[1_{\{W_{s_k} = r \}} \right] + 1_{\{ J_k = 3 \}} E_q \left[1_{\{W_{t_k} = r \}} \right]\right) \\
&+ \sum_{k = 1}^n \sum_{r = 1}^K \sum_{m = 1}^K 1_{\{ R_k = 1 \}} E_q \left[1_{\{W_{s_k} = r \}} 1_{\{W_{t_k} = m \}}\right] \log \rho_{m, r} \\
&+ \sum_{k = 1}^n \sum_{r = 1}^K \sum_{m = 1}^K 1_{\{ R_k = 0 \}} E_q \left[1_{\{W_{s_k} = r \}} 1_{\{W_{t_k} = m \}}\right] \log (1 - \rho_{m, r}),
\intertext{and employing the mean-field family assumption,}
E_q & \left[ \log p\left( W(n), (e_k)_{k = 1}^n \mid \boldsymbol{\pi}, \boldsymbol{\rho} \right) \right]\\
 =& \sum_{k = 1}^n \sum_{r = 1}^K \left( 1_{\{ J_k = 1 \}} E_q \left[1_{\{W_{s_k} = r \}} \right] + 1_{\{ J_k = 3 \}} E_q \left[1_{\{W_{t_k} = r \}} \right]\right) \\
&+ \sum_{k = 1}^n \sum_{r = 1}^K \sum_{m = 1}^K 1_{\{ R_k = 1 \}} E_q \left[1_{\{W_{s_k} = r \}} \right] E_q \left[ 1_{\{W_{t_k} = m \}}\right] \log \rho_{m, r} \\
&+ \sum_{k = 1}^n \sum_{r = 1}^K \sum_{m = 1}^K 1_{\{ R_k = 0 \}} E_q \left[1_{\{W_{s_k} = r \}} \right] E_q \left[ 1_{\{W_{t_k} = m \}}\right] \log (1 - \rho_{m, r}) \\
=& \sum_{k = 1}^n \sum_{r = 1}^K \left( 1_{\{ J_k = 1 \}} \tau_{s_k, r} + 1_{\{ J_k = 3 \}}\tau_{t_k, r} \right) \\
&+ \sum_{k = 1}^n \sum_{r = 1}^K \sum_{m = 1}^K 1_{\{ R_k = 1 \}} \tau_{s_k, r}\tau_{t_k, m} \log \rho_{m, r} \\
&+ \sum_{k = 1}^n \sum_{r = 1}^K \sum_{m = 1}^K 1_{\{ R_k = 0 \}} \tau_{s_k, r}\tau_{t_k, m} \log (1 - \rho_{m, r}).
\end{align*}
Finally, the entropy term is given by
\begin{align*}
E_q\left[\log q(W(n))\right] =& E_q\left[\sum_{v \in V(n)} \sum_{r = 1}^K 1_{\{W_v = r \}} \log \tau_{v, r}\right] \\
=& \sum_{v \in V(n)} \sum_{r = 1}^K E_q\left[ 1_{\{W_v = r \}} \right] \log \tau_{v, r} \\
=& \sum_{v \in V(n)} \sum_{r = 1}^K \tau_{v, r} \log \tau_{v, r}.
\end{align*}

\subsection*{Derivation of the E-Step}

In this section we derive the E-step of the variational EM algorithm (Step 1 of Algorithm \ref{alg:VEM}). Recall that the E-step maximizes the ELBO with respect to the variational density $q$. We perform this optimization with coordinate ascent. From \cite{blei2017variational}, for every $\ell = 1, \dots, K$, the optimal $q_w(W_w)$ satisfies
\begin{align*}
\tau_{w, \ell} = q^\star_w\left(W_w = \ell \right) \propto & \exp \left\lbrace E_{q_{-w}}\left[ \log p\left( W_w = \ell \mid (W_v)_{v \neq w}, (e_k)_{k = 1}^n, \boldsymbol{\pi}, \boldsymbol{\rho} \right) \right] \right\rbrace,
\intertext{which, by definition of conditional density, is proportional to}
\propto & \exp \left\lbrace E_{q_{-w}}\left[ \log p\left( W_w = \ell , (W_v)_{v \neq w}, (e_k)_{k = 1}^n \mid \boldsymbol{\pi}, \boldsymbol{\rho} \right) \right] \right\rbrace.
\end{align*}
Here, $q_{-w}$ denotes the variational density on $(W_v)_{v \neq w}$. Up to some constant $C$ not depending on $w$, the log-likelihood term is given by 
\begin{align*}
\log p\left( W_w = \ell , (W_v)_{v \neq w}, (e_k)_{k = 1}^n \mid \boldsymbol{\pi}, \boldsymbol{\rho} \right) =& \log \pi_{\ell} + \sum_{m = 1}^K \log \rho_{m, \ell} \sum_{k  : s_k = w} 1_{\{W_{t_k} = m\}} 1_{\{R_k = 1 \}} \\
&+ \sum_{m = 1}^K \log (1 - \rho_{m, \ell}) \sum_{k  : s_k = w} 1_{\{W_{t_k} = m\}} 1_{\{R_k = 0 \}} \\
&+ \sum_{r = 1}^K \log \rho_{\ell, r} \sum_{k  : t_k = w} 1_{\{W_{s_k} = r\}} 1_{\{R_k = 1 \}} \\
&+ \sum_{r = 1}^K \log (1 - \rho_{\ell, r}) \sum_{k  : t_k = w} 1_{\{W_{s_k} = r \}} 1_{\{R_k = 0 \}} + C,
\intertext{and taking an expectation with respect to $q_{-w}$ gives}
\log p\left( W_w = \ell , (W_v)_{v \neq w}, (e_k)_{k = 1}^n \mid \boldsymbol{\pi}, \boldsymbol{\rho} \right) =& \log \pi_{\ell} + \sum_{m = 1}^K \log \rho_{m, \ell} \sum_{k  : s_k = w} \tau_{t_k, m} 1_{\{R_k = 1 \}} \\
&+ \sum_{m = 1}^K \log (1 - \rho_{m, \ell}) \sum_{k  : s_k = w} \tau_{t_k, m} 1_{\{R_k = 0 \}} \\
&+ \sum_{r = 1}^K \log \rho_{\ell, r} \sum_{k  : t_k = w} \tau_{s_k, r} 1_{\{R_k = 1 \}} \\
&+ \sum_{r = 1}^K \log (1 - \rho_{\ell, r}) \sum_{k  : t_k = w} \tau_{s_k, r} 1_{\{R_k = 0 \}} + C.
\end{align*}
Hence, 
\begin{align}
\label{eq:update}
\begin{split}
\tau_{w, \ell} = q^\star_w\left(W_w = \ell \right) \propto & \exp \left\lbrace E_{q_{-w}}\left[ \log p\left( W_w = \ell , (W_v)_{v \neq w}, (e_k)_{k = 1}^n \mid \boldsymbol{\pi}, \boldsymbol{\rho} \right) \right] \right\rbrace \\
\propto & \pi_\ell \prod_{m = 1}^K \rho_{m, \ell}^{\sum_{k  : s_k = w} \tau_{t_k, m} 1_{\{R_k = 1 \}}} (1 - \rho_{m, \ell})^{\sum_{k  : s_k = w} \tau_{t_k, m} 1_{\{R_k = 0 \}}} \\
& \times \prod_{r = 1}^K \rho_{\ell, r}^{\sum_{k  : t_k = w} \tau_{s_k, r} 1_{\{R_k = 1 \}}} (1 - \rho_{\ell, r})^{\sum_{k  : t_k = w} \tau_{s_k, r} 1_{\{R_k = 0 \}}}.
\end{split}
\end{align}
Thus, in the E-step, one cycles through \eqref{eq:update} for each $w \in V(n)$ until the ELBO no longer meaningfully increases.

\vskip 0.2in
\bibliography{PAMultipleReciprocity}

\end{document}